\documentclass[final,3p,times]{elsarticle}
\usepackage{amssymb}
\usepackage{pifont}
\usepackage{longtable}
\usepackage{amsmath}
\usepackage{colortbl}
\usepackage{booktabs}
\usepackage{caption}
\usepackage{xcolor}
\usepackage{float}
\usepackage{bm}
\usepackage{graphicx}
\usepackage{amssymb}
\usepackage{orcidlink}
\usepackage{algorithm}
\usepackage{algpseudocode}
\usepackage{multirow}
\usepackage{colortbl}
\usepackage{graphicx}
\usepackage{longtable}
\usepackage{booktabs}
\usepackage{amsthm}
\usepackage{supertabular}

\definecolor{top1}{RGB}{255,0,0} % Red
\definecolor{top2}{RGB}{0,0,255} % Blue
\definecolor{top3}{RGB}{0,255,0} % Green
\definecolor{top4}{RGB}{128,0,128} % Green
\newcommand{\add}[1]{{\color{black}{#1}}}
\newcommand{\del}[1]{\iffalse{#1}\fi}

\newcommand{\aug}[1]{{\color{black}{#1}}}
\newcommand{\rem}[1]{\iffalse{#1}\fi}

\journal{Information Fusion}

\begin{document}

\begin{frontmatter}



\title{FS-Diff: Semantic Guidance and \del{Resolution}\add{Clarity}-Aware Simultaneous \add{Multimodal} Image Fusion and Super-Resolution}


% \author{Yushen Xu, Yuchan Jie, Xiaosong Li} %% Author name
\author[author1]{Yuchan Jie \fnref{co-first}}
\ead{jyc981214@163.com}

\author[author2]{Yushen Xu \fnref{co-first}}
\ead{2112355010@stu.fosu.edu.cn}

\author[author2]{Xiaosong Li\corref{cor}}
\ead{lixiaosong@buaa.edu.cn}

\author[author3]{Fuqiang Zhou}
\ead{zfq@buaa.edu.cn}

\author[author1]{Jianming Lv}
\ead{jmlv@scut.edu.cn}

\author[author4]{Huafeng Li}
\ead{hfchina99@163.com}

\fntext[co-first]{Equal Contribution}
\cortext[cor]{Corresponding author}

% \address[label1]{address1, China}
\address[author1]{School of Computer Science and Engineering, South China University of Technology, 510006, Guangzhou, China}

\address[author2]{ School of Physics and Optoelectronic Engineering, Foshan University, 528225, Foshan, China}

\address[author3]{ School of Instrumentation Science and Optelectronics Engineering, Beihang University, 100191, Beijing, China}

\address[author4]{ School of  Information Engineering and Automation, Kunming University of Science and Technology, 650500, Kunming, China}

%\affiliation{organization={Foshan University},%Department and Organization
%            city={Foshan},
%            postcode={528225}, 
%           state={GuangDong},
%           country={China}}

%% Abstract

\begin{abstract}
As an influential information fusion and low-level vision technique, image fusion integrates complementary information from source images to yield an informative fused image. A few attempts \rem{were}\aug{have been} made in recent years to jointly realize image fusion and super-resolution. However, in real-world applications such as military reconnaissance and
long-range detection missions, the target and background structures in multimodal images are easily corrupted, with low resolution and weak semantic information, which \rem{causes}\aug{leads to} suboptimal results in current fusion techniques. In response, we propose FS-Diff, a semantic guidance and clarity-aware joint image fusion and super-resolution method. FS-Diff unifies image fusion and super-resolution as a conditional generation problem\rem{, leveraging}\aug{. It leverages} semantic guidance from the proposed clarity sensing mechanism \rem{(CLSE) with a clarity-aware vision language model (CA-CLIP) }for adaptive low-resolution perception and cross-modal feature extraction. Specifically, we initialize the desired fused result as pure Gaussian noise and introduce the bidirectional feature Mamba to extract \aug{the} global features of the multimodal images. \rem{Meanwhile}\aug{Moreover}, utilizing the source images and semantics as conditions, we implement a random iterative denoising process \rem{by}\aug{via} a modified U-Net network\rem{, which is }\aug{. This network is }trained for denoising at multiple noise levels to produce high-resolution fusion results with cross-modal features and abundant semantic information. We also construct a
powerful aerial view multiscene (AVMS) benchmark \rem{covers}\aug{covering} 859 pairs of images. Extensive joint image fusion and super-resolution experiments on six public and our AVMS datasets demonstrated that FS-Diff \rem{excels}\aug{outperforms} the state-of-the-art methods at multiple magnifications and can recover richer details and semantics in the fused images. The code is available at https://github.com/XylonXu01/FS-Diff.
%In real-world scenarios encompassing military reconnaissance and crop detection, the object and context structures in the image are always destroyed owing to the long shooting distance, resulting in serious degradation of the image resolution and the blurring of semantic information. Owing to imaging sensors constraints, it is challenging for a single imaging device to fully capture scene information, and existing fusion methods have limited global feature modeling capabilities and cross-modal learning abilities. To tackle this issue, we propose a novel \del{resolution}\add{clarity}-aware joint multi\del{-source}\add{modal} image fusion and super-resolution method guided by semantic guidance, called FS-Diff.
\end{abstract}

% %Graphical abstract
% \includegraphics{image/01.pdf}

% \begin{highlights}
% \end{highlights}

%% Keywords

\end{frontmatter}




\section{Introduction}
Owing to the limitations of optical imaging systems, images obtained \rem{using}\aug{via} a single imaging device can only characterize part of a scene. For example,\rem{ the information of } infrared (IR) images \rem{is}\aug{are} limited by \rem{a}\aug{their} specific spectral range and imaging resolution\rem{, which}\aug{. They} can \rem{only }present \aug{only} some of the thermal radiation characteristics \rem{in}\aug{of} a target and cannot reflect the texture and details of the visible image (VI)\rem{, which is a typical case of constrained modal information}. Multimodal image fusion (MMIF) is an essential field within image processing\rem{, focusing}\aug{. It focuses} on integrating complementary information from diverse sensors to generate a fusion result with a more complete representation of the scene \cite{1,2,3,4,5,6}. Fused images with superior visual perception and excellent scene and target representation are widely used in fields such as biometrics \cite{7}, semantic segmentation \cite{8}, and target detection \cite{9}. Medical image fusion (MIF) and visible and infrared image fusion (VIRF) are challenging MMIF tasks.

MIF can simultaneously represent the metabolic and structural information \rem{of}\aug{in} functional and anatomical medical images. \rem{In}\aug{By} contrast, VIRF aims to preserve the high-spatial-resolution texture information in VI and the thermal target information in IR, \rem{which}\aug{as IR} is not sensitive to light. In the past few decades, various MMIF techniques have been developed to enhance the quality of fused images. These fusion approaches can be classified as multiscale transform (MST)-based \cite{10}, sparse representation (SR)-based \cite{11}, saliency detection (SD)-based \cite{12}, and deep learning (DL)-based \cite{13}.

In contrast to the first three conventional fusion methods, DL-based methods can learn meaningful features from extensive training data to meet the requirements of sensitivity and fast adaptation of fusion methods in complex scenes \cite{14}. Although existing MMIF can achieve satisfactory fusion outcomes in general, the following two challenges are neglected in practical applications. First, existing fusion methods suffer from poor performance in adaptively processing low-resolution (LR) source images, making it difficult to fulfill the requirements of applications such as unmanned aerial vehicle (UAV) reconnaissance, crop monitoring, and autonomous driving. In these applications, the state of the target or camera changes rapidly, \rem{which thereby places a higher demand on the amount of }\aug{increasing the demand for} image information and resolution. In particular, when UAVs \rem{shoot}\aug{capture images} at high altitudes, multiple factors, such as flight altitude, steadiness, and atmospheric conditions, significantly affect image clarity. Therefore, in unfavorable imaging environments, such as long-distance shooting, and under the performance limitations of imaging systems, it is \rem{of practical significance}\aug{crucial} to investigate methods that can adaptively perceive resolution degradation and flexibly incorporate semantic information while preserving cross-modal features\rem{, which is a challenge that urgently needs to be solved}. \aug{This is a challenge that urgently needs to be solved.}

\begin{figure*}[htbp]
\centering	
\includegraphics[width=1.0\linewidth]{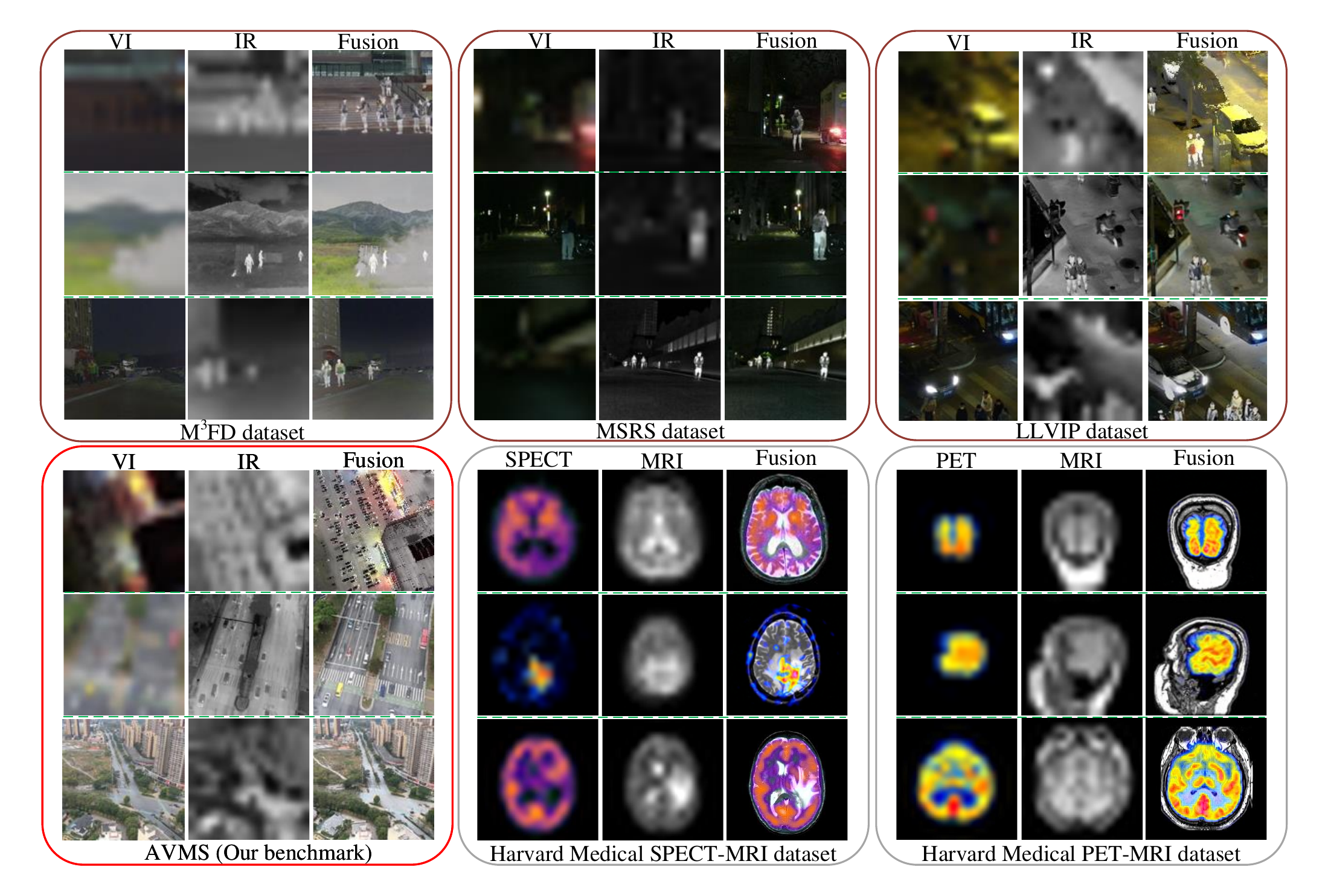}
\caption{Our FS-Diff demonstrates the outstanding fusion performance of joint image fusion and super-resolution with a scaling factor of 8 in \rem{seven}\aug{five} public and our AVMS datasets. \rem{The top four boxes show the VI-IR dataset (showing three fusion cases: both VI and IR blurred, VI blurred, IR blurred), and the bottom four boxes show the multimodal medical and multi-focus image dataset (showing both images blurred). The proposed method can produce high-resolution fusion results for any source image blurred on different fusion tasks.}\aug{The first four boxes show the VI-IR dataset, demonstrating three fusion scenarios: both VI and IR blurred, VI blurred, and IR blurred. The remaining two boxes present the multimodal medical dataset, where both images are blurred.}}
\label{Frame1}
\end{figure*}

\rem{Usually}\aug{Generally}, image fusion and super-resolution tasks can only be completed in steps\rem{,}\aug{. However, this approach}\rem{ but this manner of processing is likely to allow the errors of previous steps to be further amplified in subsequent steps, which can lead to a serious loss of modal information.} \aug{can amplify errors from previous steps in subsequent steps, leading to a serious loss of modal information.} Despite \rem{researchers attempting}\aug{efforts} to solve this problem \cite{15,16,17,100}, several challenges \aug{still} remain.\rem{ In the latest work, } Xiao et al. \cite{16} proposed a heterogeneous knowledge distillation network model\aug{. It}\rem{that} attempts to obtain high-resolution (HR) fused images by using a teacher network to direct the student network through a corner-embedded attention mechanism. Li et al.\cite{15} proposed a meta-learning-based fusion framework \rem{that can fuse}\aug{for fusing} images with different resolutions. However, their generalizability and flexibility \rem{perform poorly}\aug{are limited} on datasets of different types, failing to meet the high-magnification requirements of real scenes. When \rem{dealing with}\aug{processing} LR images with severely degraded resolution, these methods \cite{15,16,17} struggle to adaptively perceive resolution and incorporate semantic information into the fused results. They also lack robust global and cross-modal feature extraction capabilities, making the fusion results susceptible to artifacts and hindering the recovery of high-fidelity details. This is mainly because high-magnification joint image fusion and super-resolution \aug{require} not only \rem{involve }pixel value adjustment but also \rem{require }full consideration of image semantics, details, and textures.

Second, \rem{there is a lack of MMIF datasets for aerial photography. Compared with other computer vision tasks, MMIF faces the problem of a lack of benchmark variety,} \aug{despite advancements in specialized applications, existing aerial photography datasets present inherent limitations that hinder their direct adaptation to the MMIF task. For example, DroneVehicle \cite{114} provides RGB-IR image pairs in diverse lighting conditions but lacks object diversity. Anti-UAV410 \cite{115} only provides thermal infrared unimodal data and lacks visible counterparts. VEDAI \cite{117} is limited by confined lighting conditions and a narrow range of target categories.} \aug{Furthermore, although} \rem{Although} existing \rem{image fusion}\aug{MMIF} datasets \aug{including}\rem{such as} TNO \cite{18}, RoadScene \cite{19}, LLVIP \cite{20}, M\textsuperscript{3}FD \cite{21}, and MSRS \cite{22} are widely used, they are not specifically constructed for the dual tasks of simultaneous image super-resolution and fusion in long-distance shooting (\textit{e.g.}, aerial photography using a UAV). Specifically, these datasets \rem{are unable to cover a wide range of degraded-resolution types of aerial images and lack diverse shooting distances, shooting viewpoints, and a wide range of shooting scenes.}\aug{lack coverage of degraded-resolution aerial images, diverse shooting distances, varied viewpoints, and a wide range of scenes.} Additionally, \del{MSIF}\add{MMIF} has the potential to address the inconsistent performance of unimodal images in detection and segmentation tasks \cite{21}. However, owing to the lack of annotated targets in available datasets,\rem{ it is difficult for researchers to} assess\aug{ing} the effectiveness of fusion approaches based on the performance of fused images in target detection or semantic segmentation tasks \aug{is difficult for the researchers}.

Generative models are widely used in DL-based MMIF owing to their superior visual perception. \rem{Such}\aug{These} models \rem{are }effective\aug{ly} \rem{in analyzing}\aug{analyze} the empirical distributions of complex scenes through \rem{deeply examining}\aug{an in-depth examination of the} intrinsic laws and statistical properties of the images. Among these, generative adversarial networks (GANs)-based image fusion models \cite{21,23,24} are predominant. Such models typically consist of a generator that is dedicated to fusion image generation and a discriminator that evaluates the distribution similarity between the fused result and the source image. Despite the potential of GAN-based methods for generating satisfying fusion results, they experience unstable optimization, and their model-building process requires careful regularization and optimization strategies. In addition, GAN-based methods often \rem{suffer from}\aug{face} issues \rem{like}\aug{such as} mode collapse and training instability \cite{112}. As a successor to the GAN model, the denoising diffusion probabilistic model (DDPM) \cite{25} recovers superior images via diffusion and denoising processes. Compared with GAN, DDPM exhibits a steadier optimization process and stronger interpretability. Zhao et al. \cite{2} designed the fusion method utilizing DDPM\aug{, which}\rem{ that} splits the conditional generation problem into a maximum likelihood subproblem and an unconditional generation subproblem\rem{, thereby providing}\aug{. This provides} a new idea for the design of fusion methods based on generative models. \rem{In addition to}\aug{Beyond} these advancements\rem{ in generative models},\rem{ their integration} \aug{integrating generative models} with vision-language model (VLM) offers new opportunities to enhance the quality of generated images. VLM can more precisely understand and generate content that aligns with specific semantics. By combining the strengths of the generative model with VLM,\rem{there is potential to further advance} the field of MMIF \aug{can be advanced further}.

To this end, we propose a novel approach called FS-Diff\rem{, which}\aug{. It} effectively utilizes generative priors and semantics to guide the model for clarity adaptive perception and intra- and inter-modal feature extraction to jointly achieve MMIF and super-resolution. FS-Diff is adaptive \rem{to sense}\aug{for sensing} the degradation of resolution, with multi\del{-source}\add{modal} learning capabilities, and includes forward and backward Markovian diffusion processes. The forward Markovian process gradually adds Gaussian noise to the HR fusion image obtained by fusing the HR source images using CDD \cite{27}\rem{, which}\aug{. This process} efficiently processes multimodal information, captures deep correlations between modalities, and achieves good fusion results. The reverse Markovian process connects the noisy target image of the forward process with the LR source image\aug{.} \aug{It}\rem{that} constructs a joint representation of the multimodal data in the latent space by introducing a bidirectional feature Mamba (BFM) block \cite{102}\aug{,} \rem{that}\aug{which} can perceive global information to train a modified U-Net network \cite{28} to predict and eliminate the different levels of noise added by the forward process to recover the original HR fusion image. The inference phase starts with pure Gaussian noise\rem{, and the}\aug{. The} HR fusion results are obtained by iteratively refining the noise removal by using a denoising network with multiple noise removal capabilities, conditioned on multimodal LR images and semantic guidance. \aug{To achieve precise semantic embedding during fusion, we aggregate semantic extraction and clarity perception capabilities through the proposed clarity sensing (CLSE) mechanism and the clarity-aware contrastive language-image pre-training (CA-CLIP) model.} \rem{To ensure the effective embedding of semantic information during the fusion process, we aggregate the semantic extraction and clarity perception capabilities of the proposed  clarity sensing (CLSE) mechanism with Clarity-aware Contrastive Language-Image Pre-training (CA-CLIP) model based on DA-CLIP \cite{105} to provide accurate semantic guidance for the fusion process.}Through the cross attention module, deep semantics are effectively injected into \aug{the} fusion features. The combination of the generative model with BFM and CLSE   significantly enhances the ability of FS-Diff to handle cross-modal and detail feature extraction in the presence of image resolution degradation. As shown in Figure \ref{Frame1}, FS-Diff has high joint fusion and super-resolution performance with good generalization on VIRF, MIF, and multifocus tasks.

The main contributions of this paper are as follows. 

\begin{enumerate}
    %\item For the first time, we achieve end-to-end clarity-adaptive recognition and joint M\del{S}\add{M}IF and super-resolution tasks at \del{multiple} high magnification \add{($\bm{\times8}$)} under \add{reduced resolution} conditions \del{of reduced resolution} in single or dual-source images, thereby increasing the flexibility and \add{applicability} \del{range} of M\del{S}\add{M}IF \del{applications} in complex scenes.
    \item We propose an end-to-end semantic-guided and resolution-aware fusion network. To the best of our knowledge, this is the first \rem{time that}\aug{study to achieve synchronous fusion and super-resolution of pairs of LR multimodal image pairs with inconsistent clarity} \rem{automatic clarity awareness under LR multimodal images and synchronizes realize image fusion and super-resolution reconstruction tasks}at multiple high magnifications.
    
    %\item We present \add{a novel CLSE mechanism with clarity-adaptive sensing capability}\del{RA-CLIP endowed with clarity adaptive sensing capability} to extract effective semantic information from source images in advance to guide the fusion process. \del{The image controller in it can calculate the degradation rate of the input images and adjust the image encoder to output content embeddings with semantic guidance.}

  \item We \aug{introduce the}\rem{propose} CLSE mechanism, where the core component is our proposed CA-CLIP, \aug{which is} endowed with a clarity-adaptive sensing capability to extract semantic information from source images in advance to guide the fusion process. \del{The image controller can calculate the degradation rate of the input images and adjust the encoder to output content embeddings with semantic guidance.}

    \item We introduce the BFM to create a consolidated joint image representation during the fusion process to model and extract global features from multimodal images.
    
%    \item For the first time, we constructed an aerial view multiscene (AVMS) dataset with 859 aligned VI and IR image pairs as well as 3821 annotated targets that cover four types of lighting conditions, including daytime, nighttime, dusk, and complex weather as well as more than seven different scene types. This dataset is suitable for image fusion, super-resolution, long-distance detection in complex scenes, and semantic segmentation.

\item We construct an aerial view multiscene (AVMS) dataset with 859 aligned VI and IR image pairs, 3821 annotated targets that cover daytime, nighttime, dusk, and complex weather conditions, and more than seven different scenes. Our dataset is suitable for image fusion, super-resolution, long-distance detection, and semantic segmentation tasks.

\end{enumerate}

The remainder of this paper is organized as follows. Section 2 presents a review of the research background. Section 3 details the motivation and algorithmic specifics of FS-Diff. Section 4 describes the created AVMS dataset. Section 5 presents the experimental validations. Section 6 compares the computational complexity of FS-Diff with those of other methods. Section 7 concludes the paper by outlining the contributions and application prospects.

\section{Related work}
\subsection{Image fusion methods}
\subsubsection{Traditional fusion methods}
Traditional image fusion methods include MST-based, SR-based, and SD-based \rem{methods}\aug{approaches}. Among these, MST-based methods \cite{10,11} include three steps: decomposition, fusion, and reconstruction. First, typical MST tools, such as non-subsampled contourlet transforms \cite{29} and edge-preserving filters \cite{30,31}, are introduced to decompose the image into multiscale components. Subsequently, specific fusion rules are used to measure the pixel activity of the image, and pixels or regions with high activity are selected. Finally, the fused image is reconstructed using the inverse MST. MST-based methods can reduce the generation of artifacts during the fusion process owing to translation invariance. However, \rem{if}\aug{inappropriate selection of} the decomposition tool or number of decomposition layers\rem{is not properly chosen, it} will lead to a loss of information in the fused image \cite{14}.

SR achieves an efficient and accurate representation of the image by decomposing it into a linear combination of sparse coefficients and an overcomplete dictionary. SR-based methods \cite{32,33,34} assume that the activity level of an image can be computed in a sparse domain \cite{35}. They first segment the image into multiple overlapping blocks via a sliding window, vectorize them\rem{,}\aug{. Then, they} compute the sparse coefficients using sparse coding, and finally fuse them to reconstruct the fused image. Although SR-based methods have better noise robustness, balancing fusion performance and efficiency remains a challenge \cite{36}.

SD mimics the visual mechanism\rem{ of the human eye} to highlight regions of an image that are significant to the surroundings \cite{37}.\rem{ and} \aug{It} is widely used in tasks such as image compression \cite{38} and segmentation \cite{39}. SD-based methods \cite{12} use saliency extraction operators to identify the salient regions of a source image, which in turn generates a decision map to reconstruct the result. These methods can be further classified into two categories namely, salient object detection \cite{40,41,42} and salient edge detection \cite{12,43}, both of which consider the correlation between pixels in a local region. However, SD-based methods are sensitive to significance-detection algorithms and poor significance-extraction operators can lead to discontinuities at the regional junctions of the fused image.
\subsubsection{Deep learning-based methods}
In recent years, DL-based methods aimed at enhancing the perception of fused images, such as contrast and color, have emerged \cite{15,43,44,45,46,47,48,49,50,51,120,121}. These methods can learn more effective fusion strategies under the constraint of a loss function owing to their powerful nonlinear fitting performance and learning abilities \cite{52}. DL-based methods primarily include autoencoder(AE) model-based \cite{27,53,54,55,56}, convolutional neural network (CNN)-based \cite{57,58}, and generation model (GEM)-based methods \cite{59,60,61,62,63}.

The AE-based method compresses the input data into a low-dimensional feature representation through an encoder and designs fusion layer rules to extract the features\rem{, which}\aug{. These features} are then remapped back into the data space by a decoder to complete image reconstruction. DenseFuse \cite{64} is a typical AE-based method that trains a self-encoder using the MS-COCO dataset while combining addition and L1 norm strategies to obtain fused images. CDD \cite{27} uses a self-coding network with a two-branch transformer-CNN structure and combines it with feature decomposition loss to extract cross-modal information. 
\begin{figure*}[htbp]
\centering	
\includegraphics[width=1.0\linewidth]{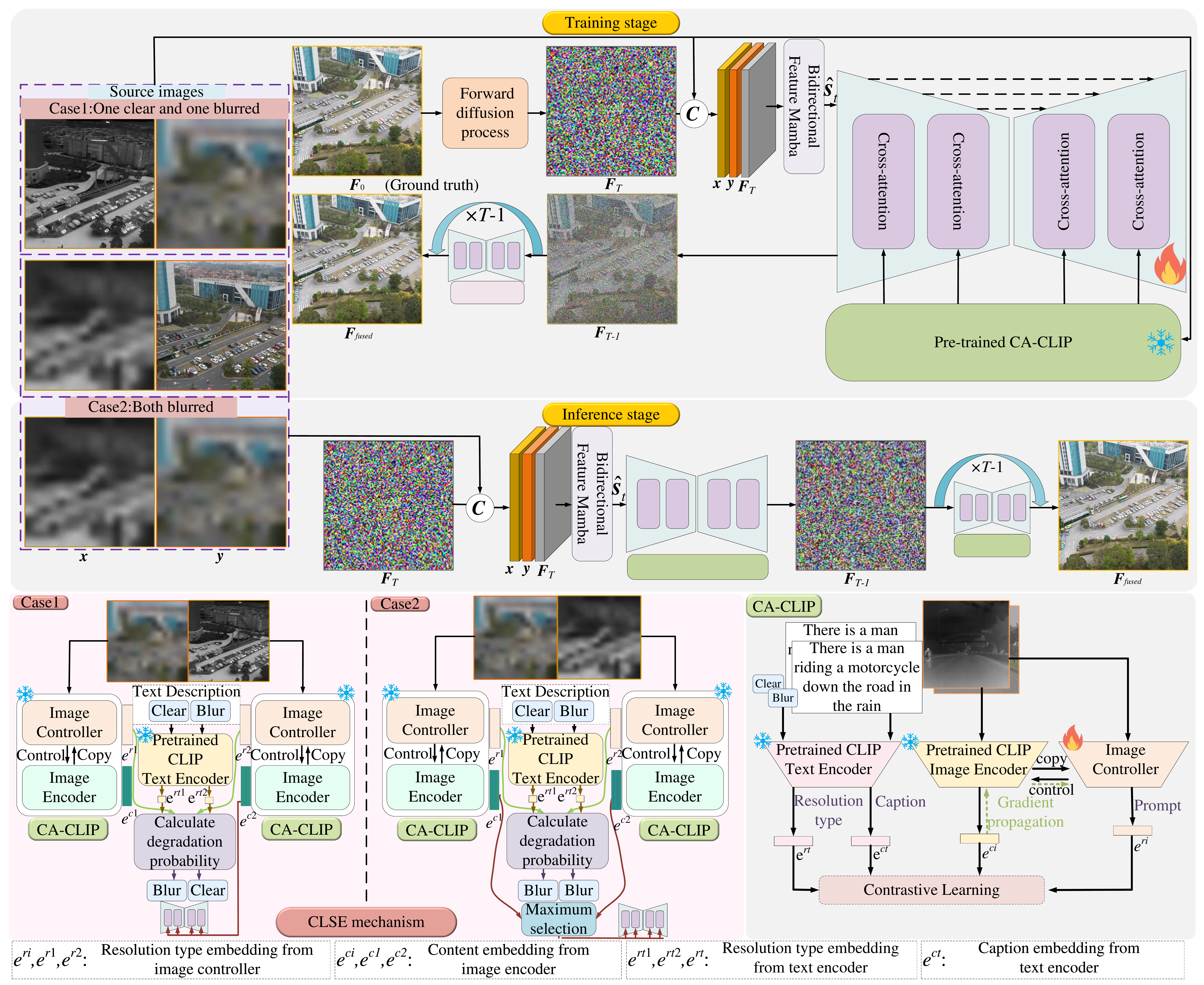}
\caption{The framework of FS-Diff. \(\bm{F}_t\) and \(\bm{F}_0\) represent the noisy output image with \textit{t} timesteps and \del{a HR fused image}\add{the ground truth}. The forward diffusion process $q$\(\left(\bm{F}_t\middle|\bm{F}_{t-1}\right)\) adds Gaussian noise to \del{the target image}\add{\(\bm{F}_0\)}, and the backward diffusion process \(p_\theta(\bm{F}_{t-1}|\bm{F}_t,\bm{x},\bm{y})\) iteratively denoise\del{the target image}. \add{The CLSE mechanism dynamically selects semantic extraction strategies for multimodal images with inconsistent resolution.} \aug{In Case 1 (single-image blur), CA-CLIP's image encoder injects the content embedding extracted from the clear image into the diffusion process. For Case 2 (dual-image blur), the model instead injects the maximum cross-image content embedding that is obtained from both blurred inputs.}} 
\label{Frame2}
\end{figure*}
CNN-based methods utilize a well-designed network structure with loss functions to complete image feature extraction, fusion, and reconstruction. PMGI \cite{65}, as a representative method, divides the network into gradient and intensity paths to maintain the ratio of texture to intensity information in the fused image, and it reuses the features in the same path. To further improve the interpretability of the algorithm, LRRNet \cite{57} achieves fusion by mathematically modelling and analyzing the fusion task and establishing a link between the optimal solution and the network that implements it.\add{ CoCoNet \cite{108} employs a coupled contrastive constraint loss and multilevel attention modules to preserve hierarchical features of different modalities, mitigating artifacts in the fused results. Text-IF \cite{109} leverages semantic text guidance to achieve degradation-aware and interactive image fusion.}

GEM-based methods, particularly GAN- and diffusion-model-based fusion methods, generate fused images that approximate the source image distribution through adversarial training and probabilistic modeling strategies, respectively. Typical methods, such as TCGAN \cite{66} combine a conditional GAN and a transformer and design a wavelet fusion module to fuse cross-domain features. DDFM \cite{2} decomposes the fusion task into two subproblems, namely, maximum likelihood and unconditional generation based on Bayesian modeling\rem{, and}\aug{. It} obtains the fused image \rem{with}\aug{using} a natural image generation prior.

In addition, to meet the demands of advanced visual tasks, scholars have explored task-oriented image fusion methods\aug{.} \rem{that}\aug{These methods} apply high-level model semantic constraints to guide the training of fusion networks \cite{21,67,68,69}. \add{To enhance fusion image quality and the accuracy of subsequent perception tasks, CAF \cite{110} introduces a perception-driven loss function, formulating fusion and perception as a joint problem. MRFS \cite{111} enhances VIRF and segmentation performance by coupling these tasks and leveraging the intrinsic consistency between visual and semantic information.} DetFusion \cite{67} guides image fusion using target information obtained from target-detection network learning. Tang et al. \cite{68} proposed combining fusion and semantic segmentation modules and increasing the semantic information obtained by the fusion module through semantic loss. Both methods \cite{67,68} guided the fusion process by using information on the target location and semantic categories to improve the quality of the fused images.

Although the aforementioned methods can achieve satisfactory fusion results, they still face significant challenges in practical applications, particularly when dealing with LR MMIF. To address this challenge, some scholars have conducted preliminary studies on practice-oriented image fusion \cite{70}. Yin et al. \cite{17} proposed the interpolation of LR source images before fusing them by using an SR-based approach. Subsequently, Li et al. \cite{15} proposed the joint learning of low-rank, sparse, and transformation dictionaries as well as learning compensation dictionaries that can reduce the loss of structure in the fused image and increase the loss of information during image super-resolution. However, this dual-task step-by-step processing method \cite{15,16} \rem{not only }has high computational complexity\aug{.} \rem{ but }\aug{It may also amplify artifacts generated in the super-resolution stage during the fusion stage}\rem{ also may lead to artifacts generated in the super-resolution stage being amplified in the fusion stage}. Moreover, this method \aug{cannot handle image fusion with high scaling factors.}\rem{is not capable of image fusion with high scaling factors.}

\subsection{Text to image models}
VLMs offer significant opportunities for improving tasks involving general text and visual representations \cite{103}. These models typically consist of a text encoder and an image encoder, which leverage contrastive learning to capture the aligned multimodal features between text and images \cite{104}. As a typical example of VLMs, CLIP \cite{104} employs image and text encoders and contrastive loss to align image-text pairs. It demonstrates robust zero-shot learning capabilities in diverse downstream tasks \rem{due to}\aug{because of} its unsupervised training method and construction of large-scale datasets. To further enhance CLIP's performance on image restoration tasks, DA-CLIP emerged. DA-CLIP \cite{105} integrates VLMs with low-level visual tasks, focusing on controlling the pretrained CLIP to predict degradation types and accordingly restore high-quality image features.  CLIP-KD \cite{118} enhances the performance of student models in zero-shot classification and cross-modal retrieval tasks through feature mimicry and interactive contrastive learning between the student and teacher encoders. Additionally, Stable Diffusion \cite{71} achieves text-controlled image generation through the integration of attention and text encoding.  By leveraging text guidance, it enables the customization of image generation, facilitating interactive control of multimodal fusion.

\subsection{Diffusion models}
Diffusion models \cite{25} have been widely used in image super-resolution \cite{71,72}, image editing \cite{73,74} and image inpainting \cite{75} owing to their powerful generative abilities. In addition, the latent features of the data learned by the diffusion model during training can be used for discriminative tasks, such as image segmentation \cite{76,77} and classification \cite{78}. The central goal is to learn to recover the structure of the training data disturbed by noise via forward and backward diffusion processes. The forward process gradually adds different levels of noise to the training data over multiple time steps, whereas the reverse process learns an iterative denoising process via U-Net to recover the original image from Gaussian noise. The inference phase begins with Gaussian noise and iteratively estimates and removes noise to obtain the target output \cite{79}. Similar to Langevin dynamics, the conditional diffusion probabilistic model (C-DDPM) \cite{26} is based on conditional inputs and transforms from a standard normal distribution to an empirical data distribution through a multistage refinement process.  DiffiT \cite{119} achieves efficient high-fidelity image generation by introducing a time-dependent multihead self-attention mechanism for fine-grained control of the denoising process. \rem{It is a promising direction to apply}\aug{The application of} diffusion models to image fusion \aug{is promising}.

% 扩散模型被广泛应用于MMIF, Dif-Fusion \cite{89} 首次将扩散模型应用于VIRF task以提高融合图像的色彩保真度。 为了提高扩散模型在MMIF中原分辨率图像特征映射效率，LFDT-Fusion \cite{123}提出潜在特征引导扩散模型用于图像融合。DRMF \cite{122}提出了退化鲁棒条件扩散模型和扩散先验组合模块以减少减轻退化对MMIF的影响。值得注意的是，
\aug{Diffusion models have been widely applied in MMIF. Dif-Fusion \cite{89} pioneered the integration of diffusion models into VIRF tasks, enhancing the color fidelity of the fused images. To optimize the efficient mapping of the original resolution image features in MMIF, LFDT-Fusion \cite{123} introduced a latent feature-guided diffusion model for image fusion. Diff-IF \cite{128} presented a fusion method based on a diffusion model and incorporating fusion knowledge priors. By decomposing the diffusion model into a conditional diffusion component and a fusion knowledge prior module, it facilitates the generation of high-fidelity fused results. Additionally, DRMF \cite{122} proposed a degradation-robust conditional diffusion model to mitigate the impact of image degradation on the MMIF. However,} current image fusion methods \aug{based on diffusion models} are ineffective under conditions of single or dual-source image resolution degradation and neglect the preservation of semantic information. Moreover, MMIF and super-resolution are deterministic image-generation processes, and the output of a well-designed model should maintain a high degree of similarity to the source images. Therefore, \rem{it is essential to utilize}\aug{utilizing} the powerful generative capabilities of DDPM combined with VLM \aug{is essential} in the fusion process. This allows the model to adaptively recognize resolution degradation and generate semantically rich fused images, thereby enhancing the accuracy of advanced visual tasks.

\subsection{Mamba}
\add{The Mamba \cite{113} model has the potential for long-sequence modeling, state space models (SSMs)-based models are motivated by continuous systems. It maps a one-dimensional sequence $x(t) \in \mathbb{R} \rightarrow y(t) \in \mathbb{R}$ through hidden states $h(t) \in \mathbb{R}^N$. The representation of the continuous system is as follows:
\begin{equation}
     h'(t) = \bm{A} h(t) + \bm{B} x(t), \quad y(t) = \bm{C} h(t)
\end{equation}
where $\bm{A} \in \mathbb{R}^{N \times N}$ is the evolution parameter, $\bm{B} \in \mathbb{R}^{N \times 1}$ and $\bm{C} \in \mathbb{R}^{1 \times N}$ are projection parameters.

To implement these systems in deep learning, the continuous parameters are transformed into discrete forms using a discretization step. A timescale parameter $\Delta$ is introduced to compute discrete versions of $\bm{A}$ and $\bm{B}$, denoted as $\overline{\bm{A}}$ and $\overline{\bm{B}}$, respectively. The Zero-Order Hold method is employed for transformation, defined as:

\begin{equation}
\overline{\bm{A}} = \exp(\Delta \bm{A}), \quad \overline{\bm{B}} = (\Delta \bm{A})^{-1} (\exp(\Delta \bm{A}) - \bm{I}) \cdot \Delta \bm{B}
\end{equation}
Using discrete parameters, the continuous equations are reformulated into the following discrete-time version:
\begin{equation}
h_t = \overline{\bm{A}} h_{t-1} + \overline{\bm{B}} x_t, \quad y_t = {\bm{C}} h_t
\end{equation}

Finally, the models output results through global convolution:

\begin{equation}
\overline{\mathbf{K}} = (\overline{\bm{C}}\overline{\bm{B}}, {\bm{C}}\overline{\bm{A}}\overline{\bm{B}}, \dots, {\bm{C}}\overline{\bm{A}}^{M-1}\overline{\bm{B}}),
\
\mathbf{y} = \mathbf{x} * \overline{\mathbf{K}}
\
\end{equation}
where $\overline{\mathbf{K}} \in \mathbb{R}^M$ represents the structured convolution kernel and $M$ denotes the length of the input sequence $\mathbf{x}$.}

\section{Method}
\subsection{Motivation}
Although existing approaches \aug{\cite{89,12,5,108,27,51,125}} generally focus on improving the network's ability to extract detailed features from multi\del{-source}\add{modal} images, the \del{MSIF}\add{MMIF} task still faces three prominent challenges. First, in UAV aerial photography and fast unmanned scenarios, the problems of image structure destruction and resolution degradation due to long distances are particularly prominent. LR images not only weaken the clarity of the background and target in the fused image but also restrict the performance of the fused image in subsequent segmentation and detection tasks, thus restricting the scope of practical applications of image fusion. Current fusion methods \aug{\cite{15, 16,17}} for LR images are only effective at low magnifications and cannot adapt to the serious degradation of image resolution,\rem{ which makes}\aug{making} it difficult to meet the performance and flexibility requirements in real scenes. Second, \rem{due}\aug{owing} to the inherent variations across modalities in multimodal images, an effective fusion method that preserves both global and semantic information is crucial. However, current fusion techniques fail to adequately address these challenges, resulting in feature loss and diminished accuracy in subsequent visual tasks. Third, dataset construction is key to implementing the \del{MSIF}\add{MMIF} method in applications. From the UAV perspective, changes in flight altitude lead to variable image field-of-view sizes, which, in turn, cause variability in target sizes. This complexity further exacerbates the challenges of visual tasks, such as image fusion, semantic segmentation, and target detection, under multiple lighting conditions. However, the current publicly available \del{MSIF}\add{MMIF} datasets are not only limited in number but \aug{are} also unavailable from aerial viewpoints, which severely limits the possibility of unmanned aerial systems realizing multi\del{-source}\add{modal} heterogeneous data fusion functions in agriculture, military, and other applications.

If the aforementioned challenges are not fully considered and effectively addressed, the performance of fusion algorithms in real applications will degrade significantly. Therefore, we propose FS-Diff. The core of FS-Diff is the use of the LR input image and the prior semantics obtained by the presented \add{CLSE with} CA-CLIP to constrain the diffusion process, which alleviates the problems of the unstable training and performance degradation of DL-based methods \rem{owing to a lack of a ground truth}. \rem{Meanwhile}\aug{Moreover}, to enhance the global and cross-modal feature perception of the fusion algorithm for multi\del{-source}\add{modal} images, we introduce the BFM \del{in FD-Diff} to construct a joint representation of multi\del{-source}\add{modal} images.

\subsection{FS-Diff}
Figure \ref{Frame2} shows the main principle of the proposed FS-Diff. The aligned input images with different \del{degradation}\add{clarity} conditions are denoted by $\bm{x} \in \mathbb{R}^{3HW}$ and $\bm{y} \in \mathbb{R}^{HW}$, respectively, and the HR noise-free fused image, \add{ \textit{e.g.} ground truth from CDD \cite{27}} is denoted by $\bm{F}_0 \in \mathbb{R}^{3HW}$, with $H$ and $W$ representing the height and width of the image, respectively. FS-Diff produces the fused \rem{image}\aug{result} \del{{$\bm{F}_0$}}\add{$\bm{F}_{fused}$} via $T$ refinement iterations from an image initiated with complete noise: $\bm{F}_T \sim \mathcal{N}(\mathbf{0}, \bm{I})$. By leveraging the BFM \cite{102} to construct a joint representation of the inputs \del{$\bm{q}_\gamma$}\rem{$\bm{q}_t$} \aug{$\bm{x}, \bm{y}$} and $\bm{F}_t$. On the basis of the studied conditional distribution $p_\theta(\bm{F}_{t-1}|\bm{F}_t, \bm{x}, \bm{y})$, FS-Diff iteratively optimizes the output to achieve the sequence output $(\bm{F}_{T-1}, \bm{F}_{T-2}, \dots, \del{\bm{F}_0}\add{\bm{F}_{fused}})$, and it eventually recovers $\del{\bm{F}_0}\add{\bm{F}_{fused}} \sim p(\bm{F}\add{_{fused}}|\bm{x}, \bm{y})$.

\begin{figure*}[htbp]

\centering	
\includegraphics[width=1.0\linewidth]{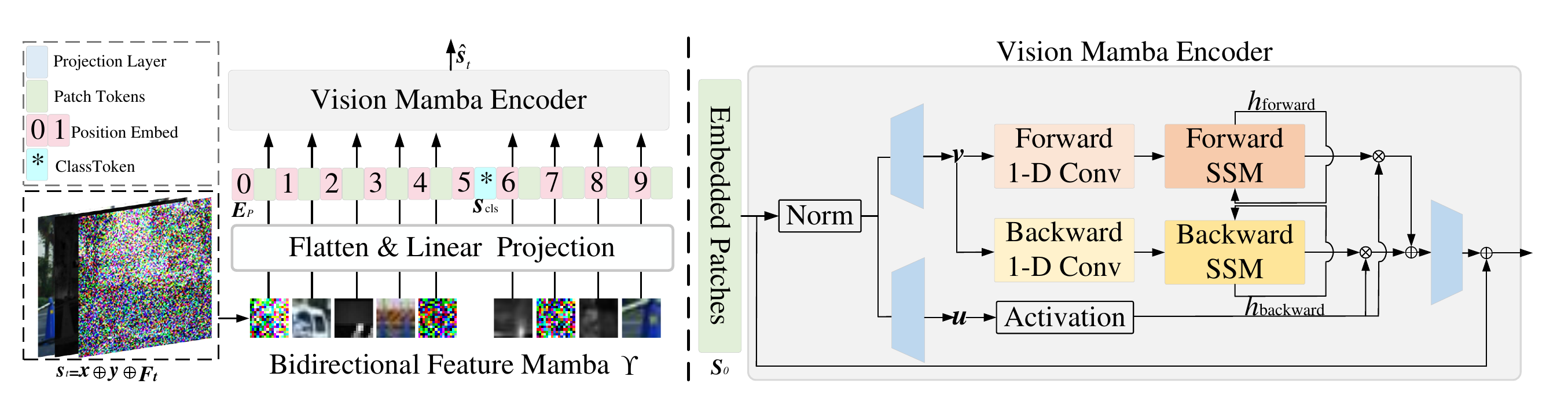}
\caption{The overview of the BFM.}
\label{FrameVim}
\vspace{-0.5cm}
\end{figure*}

During the iterative refinement process, the distribution of the intermediate images is defined by the forward diffusion process $q\left(\bm{F}_t \middle| \bm{F}_{t-1}\right)$, and noise is gradually introduced into each output image via a fixed Markov chain. The goal of FS-Diff is to remove noise based on $\bm{x}$, $\bm{y}$ and semantic guidance \add{$e^{ci}$} iteratively by using a reverse Markov chain to recover the \del{$\bm{F}_0$}\add{$\bm{F}_{fused}$} with rich semantic information and cross-modal features.
\add{
\subsubsection{Bidirectional feature Mamba}
To enhance FS-Diff's ability to extract global and information from source images, we use the BFM \cite{102} block. First, $\bm{x}$, $\bm{y}$ and $\bm{F}_t$ are concatenated along the channel dimension to form $\bm{s}_t$, which is fed into the BFM. The structure of the BFM is shown in Figure \ref{FrameVim}.

First, $\bm{s}_t \in \mathbb{R}^{H \times W \times C}$ is reshaped into patches $\bm{s}_p \in \mathbb{R}^{J \times (P^2 \cdot C)}$, where $C$, $H$, \rem{and}$W,$ \aug{$J$, and $P$} denote the number of channels, height, \rem{and}width, \aug{the number of image patches, and the size of image patches,} respectively. Then, $\bm{s}_p$ are linearly projected to the vector, and positional embeddings $\bm{E}_p$ are added:
\begin{equation}
\bm{S}_0 = [\bm{s}_{\text{cls}}; \bm{s}_p^1\bm{W}; \bm{s}_p^2\bm{W}; \dots; \bm{s}_p^j\bm{W}] + \bm{E}_p
\end{equation}
where $\bm{W} \in \mathbb{R}^{(P^2 \cdot C) \times D}$ is a learnable projection matrix, and $\bm{s}_p^j$ represents the $j$-th patch of $\bm{s}$. The class token $\bm{s}_{\text{cls}}$ denotes the whole patch sequence. 

Next, the token sequence $\bm{S}_{l-1}$ is sent to the Vision Mamba encoder \rem{that }proposed in \cite{102} to yield $\bm{S}_l$:
\begin{equation}
\bm{S}_l = \text{BFM}(\bm{S}_{l-1}) + \bm{S}_{l-1}
\end{equation}
%Here, $\text{Vim}$ represents the Vision Mamba block.
The normalization layer first normalizes $S_{l-1}$ and projects it to $\bm{v}$ and $\bm{u}$. Then, $\bm{v}$ is processed in both directions. The 1-D convolution is applied to $\bm{v}$ to obtain $\bm{s}_o^{'}$. Then, it is linearly projected into $\bm{B}_o$, $\bm{C}_o$, and $\Delta_o$. Using $\Delta_o$, $\bar{\bm{A}}_o$ and $\bar{\bm{B}}_o$ are derived. \rem{Subsequently, the output in}\aug{The outputs in the} forward and backward directions are \aug{subsequently} generated via the SSM. These outputs are then gated by $\bm{u}$ and combined to produce the final output token sequence $\bm{S}_{1}$.
%为了更简洁地描述 BFM 模块的作用，

Finally, \aug{to provide a more concise mathematical characterization of the BFM's role,} we \aug{introduce}\rem{use} \aug{$\Upsilon$ to denote the BFM and} $\hat{\bm{s}_t}$ to represent the global feature representation obtained by the BFM:
\begin{equation}
\hat{\bm{s}_t} = \Upsilon(\bm{s}_t), \quad \bm{s}_t = \bm{x} \oplus \bm{y} \oplus \bm{F}_t
\end{equation}
where \rem{$\Upsilon$ is the BMF block,} $\oplus$ represents the concatenation operation. The target image with noise is $\bm{F_t} \sim\mathcal{N}(\bm{0}, \bm{I})$.
}

\subsubsection{CA-CLIP and the CLSE mechanism}
%CA-CLIP 是基于DA-CLIP构建的,为了提升FS-Diff对原图像语义的理解和对生成高分辨率图像过程的控制。其关键是通过控制预训练的CLIP,从低分辨率原图像中获得高质量的特征，同时预测分辨率类型。见Figure2的CA-CLIP结构，image content embedding eci和clear caption embedding ect 匹配。此外，图像控制器预测的分辨率感知 embedding eri 指定分辨率类型，i.e. text encoder中对应的分辨率类型embedding ert. 
\aug{This section introduces the CA-CLIP and the CLSE mechanism. By leveraging CLIP's \cite{104} pre-trained cross-modal alignment capability, our framework enables precise semantic extraction from heterogeneous input modalities with varying clarity. Moreover, this integration enhances the generalizability of the proposed method, enabling its adaptation to more complex real-world environments.} 

Building upon DA-CLIP \cite{105}, CA-CLIP refines FS-Diff's semantic understanding of source inputs while regulating the generation process of HR fused results. Its core lies in leveraging CLIP's pretrained architecture to enable high-quality feature extraction from LR inputs. Simultaneously, it predicts the resolution type of the images (\textit{e.g.,} clear or blur), enabling the model to derive higher-quality semantic information from the clear image when only one of the images is blurred. As illustrated in Figure \ref{Frame2}, the structure of CA-CLIP involves the matching of image content embedding $e^{ci}$ and clear caption embedding $e^{ct}$. Furthermore, the resolution type embedding $e^{ri}$ predicted by the image controller specifies the resolution type, which corresponds to the resolution type embedding $e^{rt}$ from the text encoder. The image controller replicates the architecture of the image encoder, incorporating several connections initially set to zero to enhance its control capabilities and adjusting the outputs from each block of the encoder, thereby directing the encoder's predictive outputs \cite{105}.
%

%首先，训练CA-CLIP使其具备分辨率感知能力，并在低分辨率图像融合过程中冻结其权重。对于给定的文本描述（如“清晰”和“模糊”），通过预训练的CLIP文本编码器提取对应的文本特征 $e^{t1}$和$e^{t2}$，并与CA-CLIP中的图像控制器提取的特征 $e^{r1}$和$e^{r2}$ 计算分辨率退化概率，以判定哪个模态对应的原始图像为清晰图像。清晰图像的特征通过CA-CLIP的图像编码器提取，并通过cross-attention机制注入到扩散过程。如果两个模态的图像均为模糊，则选择由CA-CLIP分别获得的两个图像特征的最大值，同样通过cross-attention注入扩散过程。%
In the process of simultaneous multimodal image fusion and super-resolution, effectively harnessing semantic information is crucial. The CLIP model \cite{104}, with its pretrained capability to align cross-modal features and extract rich semantic content, plays a pivotal role in computer vision. By integrating CLIP, our framework enables the precise extraction of semantic features from multimodal images. This is achieved even when the images exhibit inconsistent clarity, ultimately aiding in the generation of HR fused images.

In our proposed CLSE mechanism, CA-CLIP is first pretrained to acquire clarity-aware capabilities. Given text descriptions, \textit{e.g.,} "clear" and "blur", the corresponding text features $e^{rt1}$ and $e^{rt2}$ are extracted by the pretrained CLIP text encoder. These features are then compared with the image features $e^{r1}$ and $e^{r2}$, extracted by the image controller, to assess the semantic similarity between the image and the given text, thereby determining which modality corresponds to the clear image. \aug{In Case 1, to increase the semantic accuracy of the fused image and reduce the potential errors introduced by the blurred image,} the feature of the clear image is extracted through the image encoder and injected into the diffusion process\rem{via the cross-attention (case1)}. \aug{In Case 2, where both images exhibit severe loss of semantic information with only luminance details preserved,}\rem{In the case that both modalities are blurred,} the most prominent of the image features obtained from the image encoder for the two images is selected and similarly injected into the diffusion process.\rem{using cross-attention (case2). } \aug{In addition, the cross-attention mechanism incorporates semantic cues into the fusion and super-resolution process. Here, the feature $\hat{s}_t$ from the BFM serves as the query vector, whereas the semantic cue $e^{ci}$ serves as both the key and value vectors. The similarity computation between the query and the key reveals regions with high similarity scores, which signify a strong correspondence between $\hat{s}_t$ and $e^{ci}$. The model then assigns higher attention weights to these regions, thereby improving the recovery of fine details.
}

\subsubsection{Diffusion process}
The forward Markovian diffusion process $q$ gradually adds Gaussian noise to \add{the ground truth} $\bm{F}_0$ over $T$ iterations \cite{25}. At time step $t$, the noisy output image $\bm{F}_t$ can be expressed as
\begin{equation}
    q\left(\bm{F}_{1:T} \middle| \bm{F}_0\right) = \prod_{t=1}^T q\left(\bm{F}_t \middle| \bm{F}_{t-1}\right)
\end{equation}
\begin{equation}
    q\left(\bm{F}_t \middle| \bm{F}_{t-1}\right) = \mathcal{N}\left(\bm{F}_t \middle| \sqrt{\alpha_t} \bm{F}_{t-1}, \left(1-\alpha_t\right) \bm{I}\right)
\end{equation}
where $\alpha_{1:T}$ ($0 < \alpha_t < 1$) represents the hyperparameters that affect the noise variance that needs to be added at each iteration, $\bm{I}$ is the unit matrix. $\bm{F}_t$ and $\bm{F}_{t-1}$ denote the outputs obtained at time steps $t$ and $t-1$, respectively.

\rem{Note that it is possible to marginalize}\aug{Notably,} the intermediate iterative process describing $\bm{F}_t$ \aug{can be marginalized} for a given $\bm{F}_0$ condition:
\begin{equation}
    q\left(\bm{F}_t \middle| \bm{F}_0\right) = \mathcal{N}\left(\bm{F}_t \middle| \sqrt{\gamma_t} \bm{F}_0, \left(1-\gamma_t\right) \bm{I}\right)
    \label{eq3}
\end{equation}
where $\gamma_t = \prod_{i=1}^T \alpha_i$, \add{$\gamma_t$ is the product of all $\alpha_i$ from the initial time step to the current time step $t$. $\bm{F}_0$ is the HR noise-free fused image}.

\subsubsection{Optimization of the denoising model}
\begin{algorithm}
\caption{Training}
\begin{algorithmic}[1]

\Repeat
    \State $(\bm{x}, \bm{y}, \bm{F}_0) \sim p(\bm{x}, \bm{y}, \bm{F})$ 
    \State $\del{\gamma}\add{\gamma_t} \sim p(\gamma)$ 
    \State $\bm{\varepsilon} \sim \mathcal{N}(\mathbf{0}, \bm{I})$ 
    \State Compute the gradient by 

    \State $\nabla_\theta \left\| f_\theta \left( \del{\bm{q}_\gamma}\add{\hat{\bm{s}}_t,} \del{\gamma}\add{\gamma_t},  \add{e^{ci}}\del{\tau_\theta (\bm{c})} \right) - \bm{\varepsilon} \right\|_2^2$ 
    
\Until{converged}
\end{algorithmic}
\end{algorithm}

The neural denoising model $f_\theta$ supplies an approximation of the gradient of the logarithmic density of the data and plays a vital role in FS-Diff. We refine a neural denoising model $f_\theta$ to predict the noise $\bm{\varepsilon}$, in each forward diffusion process and thus perceive the noise level.

To enable FS-Diff to more flexibly incorporate semantic information, we integrate a cross-attention  mechanism with U-Net of $f_\theta$. \add{The U-Net architecture consists of an encoder-decoder structure connected by a bottleneck. The encoder extracts high-level features, whereas the decoder progressively reconstructs them to the original resolution. Skip connections preserve multilevel features between the encoder and decoder. Noise-level embeddings are incorporated to handle varying noise levels. Semantic conditions are injected into the diffusion process via a cross-attention mechanism, enhancing the semantic consistency of the fused image.} \del{For one blurred image of the inputs, we introduce $\bm{c}$ to represent the semantic description of another clear image. When both images are blurred, we use the maximum value from the semantic description matrices of the two as $\bm{c}$. The CA-CLIP is denoted as $\tau_\theta$.\del{
We present the CA-CLIP, denoted as $\tau_\theta$, which based on DA-CLIP \cite{105} for better performance of FS-Diff in low-level vision tasks.} The $\bm{c}$ is projected into an intermediate representation $\tau_\theta (\bm{c})$, and cross attention is mapped to the intermediate layers of U-Net. The attention mechanism is defined as: 
$\text{Attention}(Q, K, V) = \text{softmax}\left( \frac{QK^T}{\sqrt{d}} \right) V,$ where $Q = w_Q^{(i)}(\bm{q}_\gamma), K = w_K^{(i)} \cdot \tau_\theta (\bm{c}), V = w_V^{(i)} \cdot \tau_\theta (\bm{c})$. The $d$ represents the scaling factor, and $w_Q^{(i)}, \quad w_K^{(i)}, \quad w_V^{(i)}$ are learnable mapping matrices.
\del{, $\phi_i(\bm{q}_\gamma)$ denotes the intermediate representation of U-Net.}}

%As shown in Figure \ref{Frame2}, the backbone of the denoising model $f_\theta$ uses the U-Net network from C-DDPM \cite{26}. The structure consists of a downsampling module, an intermediate connection module, and an upsampling module. The down- and up-sampling processes were implemented using four 2D convolutional layers and four transposed convolutional layers. Unlike U-Net in DDPM \cite{25}, C-DDPM replaces the residual blocks in BigGAN \cite{80} with those in DDPM and adds a self-attention mechanism. In addition, the number of residual blocks is increased, and the scale of the jump connections is modified to $\frac{1}{\sqrt{2}}$.
\del{
To describe the forward process noise variation more clearly, the HR image is expressed as
\begin{equation}
    \widetilde{\bm{F}} = \sqrt{\gamma} \bm{F}_0 + \sqrt{1 - \gamma} \bm{\varepsilon}, \quad \bm{\varepsilon} \sim \mathcal{N}(\mathbf{0}, \bm{I})
    \label{eq4}
\end{equation}
The definition of \(\widetilde{\bm{F}}\) aligns with Eq.\ref{eq3}.
}
\del{The $f_\theta(\bm{q}_\gamma, \gamma,\tau_\theta (\bm{c}))$}\add{In addition, $f_\theta(\hat{\bm{s}}_t,\gamma_t,{e}^{ci})$} contains three inputs: the joint global representation \del{$\bm{q}_\gamma$}\add{$\hat{\bm{s}}_t$}, \del{$\bm{q}_\gamma=\Upsilon (\bm{x}, \bm{y} ,\widetilde{\bm{F}})$ enhance adaptation to different modal image domains} the noise variance \del{$\gamma$}\add{$\gamma_t$} and \add{the image content embedding $e^{ci}$}\del{$\tau_\theta (\bm{c})$}.\del{Here, the $\gamma$ is generated by processing timestep $t$ using the feature-wise affine module, $\Upsilon$ is the BFM.} Algorithm 1 shows the total training process, the Kaiming initialization method \cite{81} is used to initialize the conditional noise predictor $f_\theta$. \aug{The training process begins by sampling the LR image pairs $\bm{x}, \bm{y}$ and the initial HR target image $\bm{F}_0$ from the joint distribution $p(\bm{x}, \bm{y},\bm{F})$. \rem{Subsequently, t}\aug{T}he noise intensity  $\gamma_t$ is \aug{subsequently} randomly sampled from the probability distribution $p(\gamma)$, followed by sampling noise $\bm{\varepsilon}$ from the standard normal distribution. The gradient is then computed based on the $L_2$ error between the predicted noise from $f_\theta$ and the true noise $\bm{\varepsilon}$. This iterative process continues until convergence is achieved.}

\add{The object function of FS-Diff is denoted as follows:}\del{follows the objective function of the C-DDPM, which is denoted as follows:}
\begin{equation}  
\del{    L = \mathbb{E}_{\Upsilon(\bm{x},\bm{y},\widetilde{\bm{F}}), \bm{c}, \bm{\varepsilon}, \gamma} \left[ \left\| f_\theta (\bm{q}_\gamma, \gamma,e^{ci}) - \bm{\varepsilon} \right\|_2^2 \right]}
\add{L = \mathbb{E}_{\Upsilon (\bm{x},\bm{y},\bm{F}_t), \varepsilon, t} \left[ \left\| f_\theta(\hat{\bm{s}_t},\gamma_t,{e}^{ci}) - \bm{\varepsilon} \right\|_2^2 \right]}
\end{equation}
\begin{equation}
\add{
e^{ci} = 
\begin{cases} 
e^{c1}, & \text{if } C_{VI} \land B_{IR}, \\
e^{c2}, & \text{if } C_{IR} \land B_{VI}, \\
\max(e^{c1}, e^{c2}), & \text{if } B_{VI} \land B_{IR},
\end{cases}
}
\end{equation}
where \rem{the predicted noise} $\bm{\varepsilon} \sim \mathcal{N}(\mathbf{0}, \bm{I})$. \aug{It is added to the HR target image to generate a noisy version, and $f_\theta$ is trained to predict this noise.} \rem{and} $\mathbb{E}$ represents the expected value. \del{Meanwhile, both the denoising model $f_\theta$ and $\tau_\theta (\bm{c})$ are jointly optimazed.}\add{$B_{VI}$ and $B_{IR}$ denote blurred VI and IR, respectively, whereas $C_{VI}$ and $C_{IR}$ represent clear VI and IR, respectively. $e^{c1}$ and $e^{c2}$ represent the features of VI and IR, respectively, from the image encoder of a pretrained CA-CLIP. $\land$ denotes that the conditions are satisfied simultaneously.}

\subsubsection{Inference}
\begin{algorithm}
\caption{The inference of FS-Diff}
\begin{algorithmic}[1]
\State $\bm{F}_T \sim \mathcal{N}(\mathbf{0}, \bm{I})$  
\For{$t = T, T-1, \ldots, 1$}
    \If{$t > 1$}
        \State $\bm{z} \sim \mathcal{N}(\mathbf{0}, \bm{I})$  
    \Else
        \State $\bm{z} = \bm{0}$  
    \EndIf
    \State $\bm{F}_{t-1}=\frac{1}{\sqrt{\alpha_t}}\left(\bm{F}_t - \frac{1-\alpha_t}{\sqrt{1-\gamma_t}} f_\theta(\del{\bm{q}_\gamma}\add{\hat{\bm{s}}_t,} \gamma_t, e^{ci})\del{\tau_\theta (\bm{c}))} \right) + \sqrt{1-\alpha_t} \bm{z}$  
\EndFor
\State \Return $\bm{F}_{fused}$  
\end{algorithmic}
\end{algorithm}

The inference process of FS-Diff is defined as a reverse Markovian process, as opposed to a forward diffusion process. Our goal was to model the distribution of \add{the fused result $\bm{F}_{fused}$}\del{$\bm{F}_0$} for the given $\bm{x}$ and $\bm{y}$ conditions. That is, ${p(\del{\bm{F}_0}}\add{\bm{F}_{fused}} \mid \bm{x}, \bm{y})$. The detailed computational procedure for the inference is described in Algorithm 2, which starts with the Gaussian noise $\bm{F}_T$, where $p(\bm{F}_T) = \mathcal{N}(\bm{F}_T \mid \bm{0}, \bm{I})$. \aug{For each time step $t$ (from $T$ to 1), if $t>1$, a random noise vector $\bm{z}$ is sampled from $\mathcal{N}(\mathbf{0}, \bm{I})$. When $t$=1, $\bm{z}$ is set to 0. Same as in \cite{25}, the vector $\bm{z}$ represents the noise introduced during the inference process. \rem{Subsequently, }$\bm{\varepsilon}$ is \aug{subsequently} estimated via $f_\theta$, and $\bm{F}_{t}$ is iteratively refined. Ultimately, this iterative process yields the final HR fused image, $\bm{F}_{fused}$.}

\rem{, where $\bm{z}$ denotes the noise predicted by the network}. Each iteration generates output images $\bm{F}_t \rightarrow \bm{F}_{t-1}$ with different noise levels, and as $t$ decreases, the noise level decreases. At the end of the iteration, \del{$\bm{F}_0$}\add{$\bm{F}_{fused}$} is returned as the final clear fusion result, the mathematical expression is as follows:
%\aug{For $\bm{t}>1$, $\bm{z} \sim \mathcal{N}(\mathbf{0}, \bm{I})$ serves as a random perturbation, aiding the model in exploring diverse image features and preventing it from converging to local optima. When $\bm{t}=1$, $\bm{z}$ is set to 0 to ensure the stability of the final output.} 
\begin{equation}
p_\theta(\bm{F}_{0:T} \mid \bm{x}, \bm{y}) = p(\bm{F}_T) \prod_{t=1}^T p_\theta(\bm{F}_{t-1} \mid \bm{F}_t, \bm{x}, \bm{y})
\end{equation}
\begin{equation}
p_\theta(\bm{F}_{t-1} \mid \bm{F}_t, \bm{x}, \bm{y}) = \mathcal{N}(\bm{F}_{t-1} \mid \mu_\theta(\del{\bm{x}, \bm{y}, \bm{F}_t}\add{\hat{\bm{s}_t}}, \gamma_t\add{,e^{ci}}), \sigma_t^2 \bm{I}) \label{eq:conditional_distribution}
\end{equation}

\begin{equation}
\sigma_t^2 = \frac{(1 - \gamma_{t-1})(1 - \alpha_t)}{1 - \gamma_t} \label{eq:sigma}
\end{equation}

\begin{equation}
\mu_\theta(\del{\bm{x}, \bm{y}, \bm{F}_t}\add{\hat{\bm{s}_t}}, \gamma_t\add{,e^{ci}}) = \frac{1}{\sqrt{\alpha_t}}\left(\bm{F}_t - \frac{1 - \alpha_t}{\sqrt{1 - \gamma_t}} f_\theta(\del{\bm{q}_\gamma}\add{\hat{\bm{s}_t},}\gamma_t, \del{\tau_\theta (\bm{c})}\add{e^{ci}})\right) \label{eq:mu_theta}
\end{equation}
where $\sigma_t^2$ and $\mu_\theta(\del{\bm{x}, \bm{y}, \bm{F}_t}\add{\hat{\bm{s}_t}}, \gamma_t\add{,e^{ci}})$ represent the variance and the mean of $p_\theta(\bm{F}_{t-1} \mid \bm{F}_t, \bm{x}, \bm{y})$, respectively.

\del{Given an arbitrary noisy image $\widetilde{\bm{F}}$ (containing $\bm{F}_t$),}\add{The} $f_\theta(\cdot)$ is trained to predict $\bm{\varepsilon}$. \del{Therefore,}\del{when combined with Eq.\ref{eq4}, $\bm{F}_0$}\add{$\bm{F}_{fused}$} can be approximated under the given condition for $\bm{F}_t$.\
\begin{equation}
    \del{\hat{\bm{F}}_0}\add{\hat{\bm{F}}_{fused}}= \frac{1}{\sqrt{\gamma_t}}\left(\bm{F}_t - \sqrt{1 - \gamma_t} f_\theta(\del{\bm{q}_\gamma}\add{\hat{\bm{s}_t}}, \gamma_t, \del{\tau_\theta (\bm{c})}\add{e^{ci}}))\right)
\end{equation}
where \del{\({\hat{\bm{F}}}_0\)}\add{\({\hat{\bm{F}}}_{fused}\)} is the approximation of \del{\(\bm{F}_0\)}\add{\(\bm{F}_{fused}\)}.

Finally, the refinement steps for each iteration are as follows:
\begin{equation}
    \bm{F}_{t-1} \gets \frac{1}{\sqrt{\alpha_t}} \left(\bm{F}_t - \frac{1 - \alpha_t}{\sqrt{1 - \gamma_t}} f_\theta(\del{\bm{q}_\gamma}\add{\hat{\bm{s}_t}}, \gamma_t, \del{\tau_\theta (\bm{c})}\add{e^{ci}})\right) + \sqrt{1 - \alpha_t} \bm{\varepsilon}_t
\end{equation}
where $\bm{\varepsilon}_t \sim \mathcal{N}(\mathbf{0}, \bm{I})$.
\section{Multimodality image fusion benchmark}
\begin{figure*}[htbp]
\centering	
\includegraphics[width=1.0\linewidth]{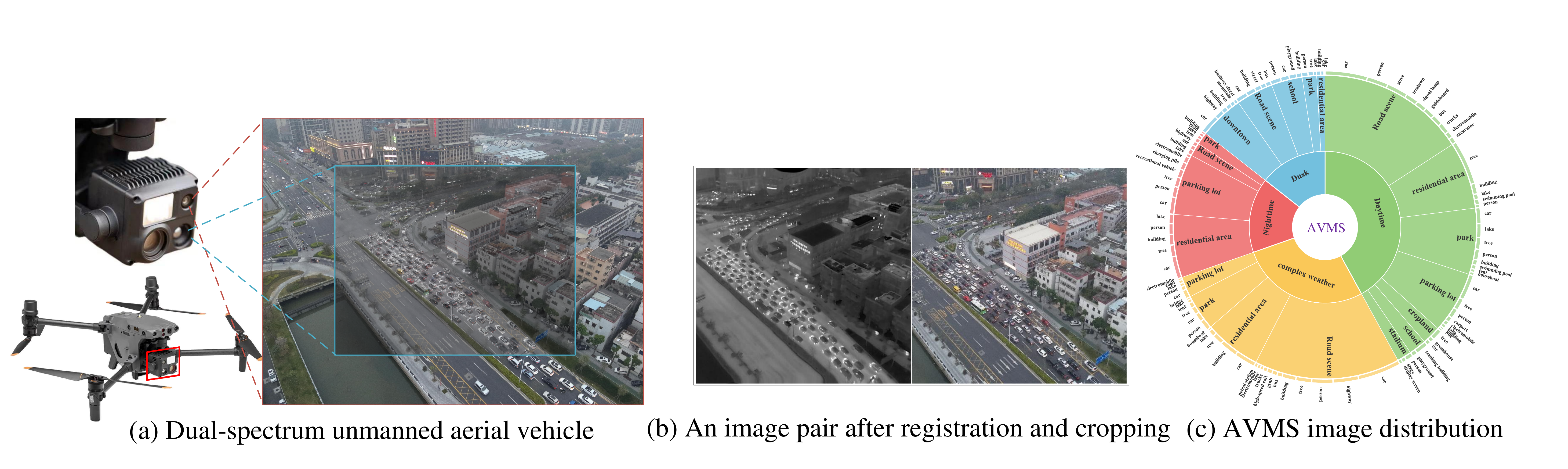}
\caption{(a) Image pair acquisition, (b) post-processing, and (c) scene distribution for AVMS. }
\label{Frame3}
\end{figure*}
\begin{figure*}[htbp]
\centering	
\includegraphics[width=1.0\linewidth]{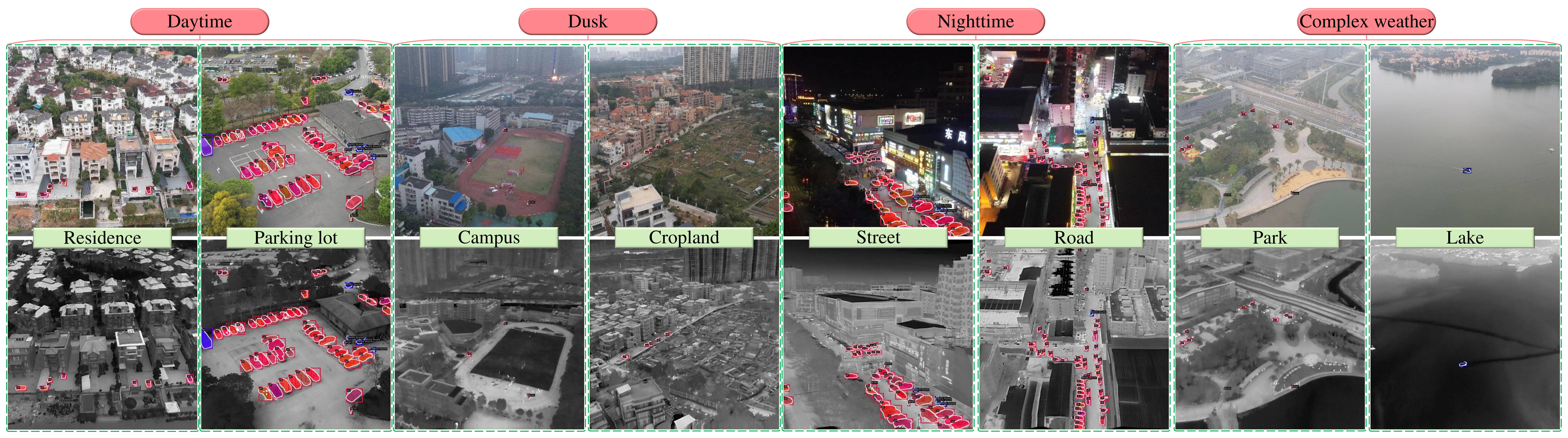}
\caption{Viewable distribution of scenes in our AVMS dataset. The first line is the visible images, and the second line is the corresponding infrared images. The text in the green box represents the scene type.}
\label{Frame4}
\end{figure*}
Multimodal datasets are crucial for building accurate and efficient deep learning-based MMIF systems \cite{20}. However, most existing multimodal image datasets \cite{18,19,20,21,22} mainly rely on handheld IR or VI cameras, which have a relatively limited image field of view. In real-world complex environments, such as battlefield \aug{environments}, UAVs need to adjust the shooting distance and camera parameters in real time to cope with changing terrain and weather, often resulting in the loss of semantic information and local structural features. This significantly affects the accuracy of advanced vision tasks.
\add{In recent years, several aerial datasets have been proposed for specific tasks, such as vehicle detection \cite{114}, drone tracking \cite{115}, and broader object detection tasks \cite{116}\cite{117}. Anti-UAV410 \cite{115} provides a thermal IR UAV tracking benchmark covering lakes, forests, and mountains, offering valuable support for anti-UAV tracking algorithm development. However, it is limited to IR images and does not support multimodal image fusion. VEDAI \cite{117} focuses on small vehicle detection from an aerial perspective, including both large and small-sized IR and VI image pairs. Nonetheless, its limited illumination conditions and few nighttime scenes restrict its application in low-light scenarios. DVTOD \cite{116} introduces a misaligned visible-thermal dataset with 54 scenes and 16 challenging attributes, but the misalignment adds complexity to the multimodal fusion algorithm design. DroneVehicle \cite{114} provides a drone-based RGB-IR image pair for vehicle detection under varying lighting conditions from day to night, yet its narrow target category limits its use in broader tasks. 

Although these datasets contribute significantly to object detection and tracking, they \rem{fall short in supporting}\aug{fail to support} the comprehensive training and evaluation of multimodal fusion algorithms} \del{To this end}\add{To address these challenges}, we established the AVMS dataset, which has a wide range of shooting angles, flexible shooting angles, and diverse lighting conditions. Unlike existing datasets, AVMS incorporates various scenes with multiple shooting distances, providing data support for developing practical deep-learning-based \del{MSIF}\add{MMIF} and super-resolution methods. AVMS is expected to be embedded in unmanned aerial systems with multisource heterogeneous data fusion functions in the future, which will provide key technical support for intelligent battlefields and UAV crop inspection.

%DJI Matrice M30T
A professional UAV was used for image acquisition, as shown in Figure \ref{Frame3}(a)). The UAV can fly and shoot steadily in windy, snowy, rainy, and low-light conditions to obtain the correct target size, and the drone can be controlled at an altitude of 5--20 m. %The infrared camera has a wavelength range of 8--14 $\mu$m. The optical center distance between the infrared and visible cameras was 3 cm, the focal lengths were 9.1 mm and 21 mm, the aperture parameters were fixed \(f/1.0\) and variable \(f/2.8\)--\(f/4.2\), and 
The standard resolutions of the captured images were 1280$\times$1024 and 4000$\times$3000. To address the issue of the inconsistent field-of-view size of the built-in camera of the UAV, we cropped and registered the VI and IR images, as shown in Figure \ref{Frame3}(b). In addition, to obtain aligned image pairs with \del{various types of degradation}\add{different clarity}, the two modal images were downsampled via bicubic interpolation.

%To solve the problem of the inconsistent field-of-view size of the built-in camera of the UAV, we first cropped the VI to have the same field of view and size as the IR. Subsequently, a symmetric bidirectional alignment method \cite{6} was used to register the IR and VI image pairs. Finally, the two modal images were down-sampled using the bicubic interpolation method to achieve aligned image pairs with multiple degeneracy types, as shown in Figure \ref{Frame3}(b).

The 859 pairs of images contained in the AVMS dataset were classified into four categories on the basis of illumination levels: daytime, dusk, nighttime, and complex weather. Figure \ref{Frame4} shows eight subscenes: residences, roads, campuses, parks, streets, stadiums, parking lots, croplands, and diverse environments in both urban and rural areas. Notably, MSIF aims to improve the accuracy of advanced vision tasks, such as detection and segmentation, to increase the effectiveness of applications in real-world scenarios, such as autonomous driving and drone combat. To this end, the AVMS dataset provides 3,821 annotated targets, including 3,407 cars, 60 buses, 3 grabs, 63 trucks, 42 electric vehicles, 188 people, 56 tents, and 2 boats, laying a critical data foundation for subsequent target detection and semantic segmentation tasks.

\section{Experiments}
This section is organized as follows. Subsection \ref{5.1} details the experimental setup. Subsections \ref{5.2} and \ref{5.3} outline comparative experiments with state-of-the-art methods on the VIRF and MIF tasks, respectively, and provide quantitative versus qualitative analyses. Subsection \ref{5.4} explores the validity of the BFM block and \del{CA-CLIP}\add{CLSE mechanism} via ablation experiments. Subsection \ref{5.5} evaluates the performance of the model on an advanced visual task. Finally, Subsection \ref{5.6} outlines extended FS-Diff experiments on multifocus image fusion.
\subsection{Experimental setups}
\label{5.1}
\textbf{1) Datasets:} Our experiments were conducted using four VIRF datasets: LLVIP (training: 12025, test: 3463) \cite{20}, M\textsuperscript{3}FD (training: 3780, test: 300) \cite{21}, MSRS (training: 1083, test: 362) \cite{22}, and our AVMS (training: 689 ,test: 170 ). MRI-PET (training: 287, test: 32) and MRI-SPECT (training: 497, test: 55) of the Harvard medical image dataset \cite{summers2003harvard} were selected for the MIF experiments.

\textbf{2) Image processing:} The HR fusion image was obtained via CDD \cite{27}, and the LR source image was down-sampled from the HR source image via a bicubic \aug{approach} to synthesize the “LR multimodal image-HR fusion image” training set. For the test set, similar to the configuration in \cite{84,85,86}, we cropped the three public datasets to $16 \times 16$, degraded them bicubically, and named them LLVIP-val, M\textsuperscript{3}FD-val, and MSRS-val.

\textbf{3) Compared methods: } We compared FS-Diff with 10 state-of-the-art (SOTA) image fusion approaches whose codes are publicly available in a VIRF task. We classified them into five groups. The first group is AE-based methods, and includes CDD \cite{27}, UMF\cite{51} \aug{and PromptFusion \cite{126}}. The second group is CNN-based methods and consists of \aug{CoCoNet \cite{108},} Reconet \cite{50}, IGNet \cite{87}, and LRRNet \cite{57}. The third group is GAN-based methods and includes Tardal \cite{21} and TGFuse \cite{88}. The fourth group is diffusion model-based methods and includes DIF \cite{89} and DDFM \cite{2}. The fifth group is  MetaFusion \cite{15}, a SOTA joint image super-resolution and fusion method. \rem{Meanwhile}\aug{Moreover}, we compared FS-Diff with SOTA methods, including CDD \cite{27}, BDLFusion \cite{45}, U2Fusion \cite{19}, and DDFM \cite{2} which are applicable to the MIF task, to prove its effectiveness in MIF.

\textbf{4) Metrics:} To objectively and comprehensively evaluate the fused results, visual information fidelity (VIF) \cite{90}, Q\textsuperscript{AB/F} \cite{91}, structural similarity index measure (SSIM) \cite{92}, peak signal to noise ratio (PSNR), learned perceptual image patch similarity (LPIPS) \cite{93}, and mean squared error (MSE) were used. The VIF, SSIM, and PSNR are image fidelity measures, whereas Q\textsuperscript{AB/F} is evaluated by quantifying the edge information, LPIPS measures the perceived quality of the image, and MSE \rem{responds}\aug{corresponds} to the error in predicting the image.

\textbf{5) Implementation details:} \aug{Our model was trained on four NVIDIA GeForce RTX 3090 GPUs and tested on a single RTX 3090 GPU.}\rem{The experiments were conducted on a machine with four NVIDIA GeForce RTX 3090 GPUs.} \aug{During training, we considered various scaling factors,}\rem{Our model was trained across various scaling factors, for} including $\bm{\times2}$, $\bm{\times4}$, and $\bm{\times8}$, for a total of 800,000 training steps with a batch size of 80. The Adam optimizer was adopted with a fixed learning rate of $1 \times 10^{-4}$. \rem{We set $T=4000$ for the inference stage.}\aug{Adapting the parameter configuration proposed in \cite{129}, we employ $T=4000$ to maintain an optimal trade-off between fusion accuracy and inference speed.} 
%为在模型性能与计算成本间达成平衡，参照文献 \ref{129},we set $T=4000$.

With the exception of MetaFusion \cite{15}, other comparison methods cannot be directly applied to the joint task of fusion and super-resolution. Therefore, these methods perform super-resolution followed by fusion in a stepwise manner for comparison with FS-Diff. \aug{Among various image super-resolution methods\cite{26,127}, given the superior performance of SR3 across multiple scaling factors, we selected it as the baseline for super-resolution.} Specifically, LR source image pairs were initially super-resolved to obtain HR images via SR3 \cite{26}, which \rem{has been}\aug{was} retrained on MMIF images, and then these HR images \rem{are}\aug{were} fused to obtain the final fused image.
%\begin{figure*}[htbp]
%\centering	
%\includegraphics[width=1.0\linewidth]{image/05.pdf}
%\caption{Fusion and super-resolution (scale:8) results for different methods on LLVIP, M\textsuperscript{3}FD, MSRS, and our AVMS datasets.}
%\label{Frame5}
%\end{figure*}
\subsection{IR and VI image fusion}
\label{5.2}
\subsubsection{Subjective assessment}
\begin{figure*}[htbp]
\centering	
\includegraphics[width=1.0\linewidth]{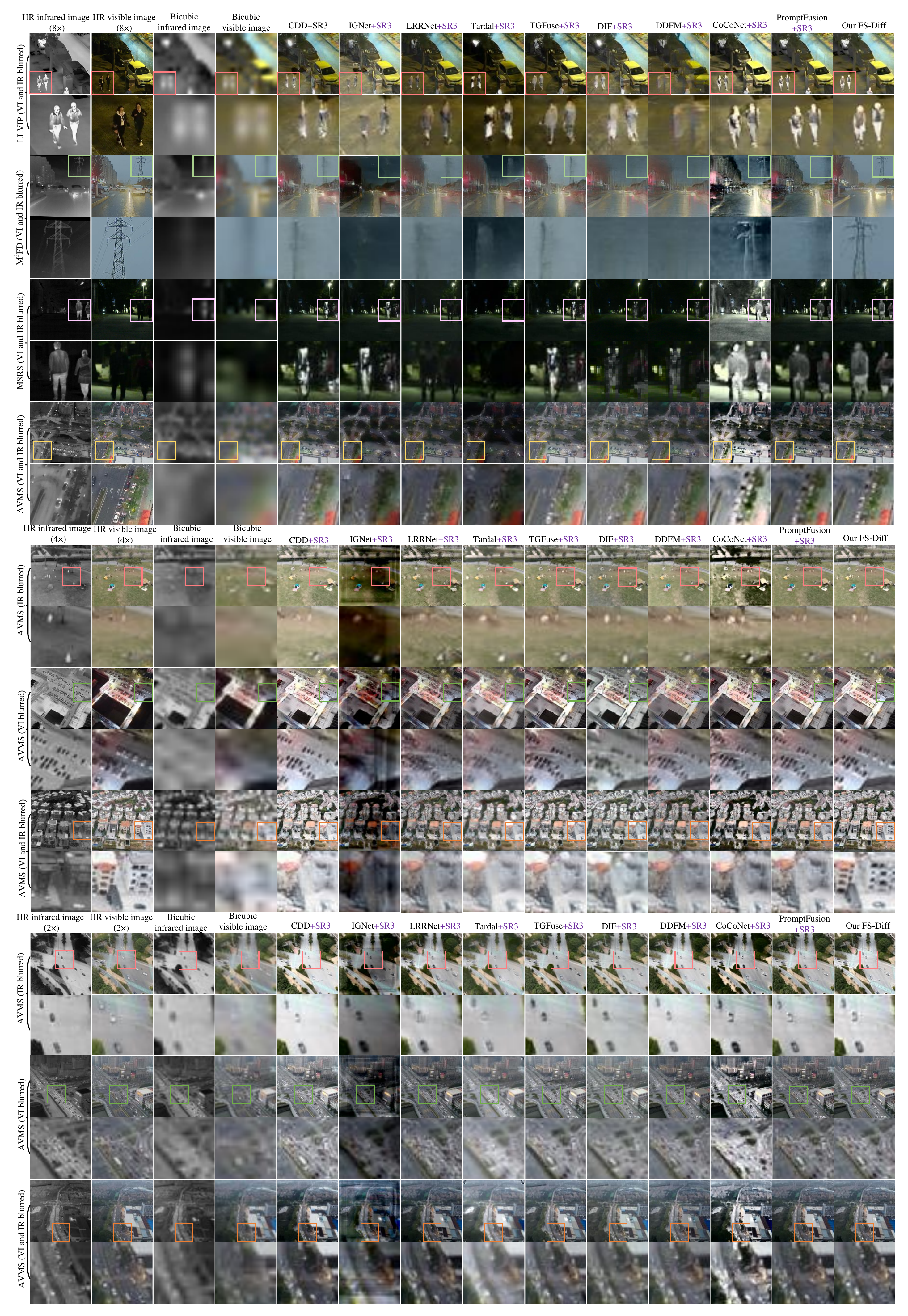}
\caption{Fusion and super-resolution (scale:8, 4 and 2) results for different methods on \aug{LLVIP, M\textsuperscript{3}FD, MSRS and} our AVMS datasets.}
\label{Frame5-842t}
\end{figure*}

Figure \ref{Frame5-842t} shows the fusion and super-resolution results of several SOTA methods with relatively better visual performance at the $\bm{\times8}$, $\bm{\times4}$, and $\bm{\times2}$ scales\rem{, generated using both the proposed method and the compared methods}. As seen from the enlarged area\del{of the Figure \ref{Frame5}}, the compared methods exhibit a degree of blurriness and a deficiency in fine-grained details and fail to retain the structures of "person" in LLVIP and MSRS, "electrical pole" in M\textsuperscript{3}FD and "cars" in AVMS.

The fusion results of IGNet+SR3, DIF+SR3, and DDFM+SR3 on the LLVIP dataset exhibit significant semantic errors in reconstructing "person" and "car". \aug{Compared with the other methods, CoCoNet+SR3 and PromptFusion+SR3 exhibit good detail preservation of "person" in the LLVIP and MSRS datasets.}\rem{In contrast, the well-trained BFM and CLSE mechanism mitigate the loss of global features in the fused images, thereby facilitating FS-Diff in generating crisp, high-resolution fusion results with realistic details and minimal distortions.} Moreover, CDD+SR3, IGNet+SR3, Tardal+SR3, \aug{and CoCoNet+SR3} demonstrate artifacts in smooth areas, such as roadways in the M\textsuperscript{3}FD dataset. \rem{MetaFusion struggles to recover semantic information from multimodal images experiencing severe resolution degradation. }Conversely, FS-Diff demonstrates greater stability in smooth areas and preserves richer contrast and significant information \rem{compared to Tardal+SR3 and IGNet+SR3}, highlighting its notable advantages in various fusion and super-resolution scenarios.

\subsubsection{Objective assessment}
Tables \ref{tab:1} and \ref{tab:add} show the VIF, Q\textsuperscript{AB/F}, SSIM, PSNR, LPIPS, and MSE values for VIRF and super-resolution on three public datasets and our constructed datasets, under three different resolution degradation scenarios of the source images. The following observations were made: (1) CDD+SR3 and \rem{DDFM}\aug{PromptFusion+SR3} demonstrate strong performance in terms of VIF and PSNR, effectively preserving visual information fidelity. Reconet+SR3 and \aug{DDFM+SR3} \rem{excel in MSE, though they lag behind UMF+SR3}\aug{outperform UMF+SR3 in terms of the MSE}. On the LLVIP dataset, LRRNet+SR3 achieves commendable PSNR scores. In the context of blurred IR images at a $\times4$ scaling factor, TGFuse+SR3 \aug{and PromptFusion+SR3} show robust overall performance. By contrast, DIF+SR3, IGNet+SR3, and Tardal+SR3 yield relatively average results. (2) As a SOTA joint image fusion and super-resolution method, MetaFusion is not capable of efficiently fusing images with severely degraded resolution, and performs poorly on six metrics. (3) FS-Diff performed well for the VIF, Q\textsuperscript{AB/F}, SSIM, and LPIPS metrics. Overall, FS-Diff, which integrates both image fusion and super-resolution tasks in a unified framework, demonstrates superior performance across all four datasets compared with methods that process these tasks separately. This fully demonstrates its ability to generate high-resolution fused images that are consistent with the visual perception of the human eye, with a more complete retention of details and semantic information.

Subjective and objective evaluations simultaneously affirmed the applicability of FS-Diff for image fusion and super-resolution at different magnifications in different scenes. This is primarily because FS-Diff is trained based on joint multitask learning, including fusion and super-resolution, which strengthens the feature recognition ability of the model.
% \usepackage{multirow}
% \usepackage{graphicx}

\begin{table*}[]
\centering
\caption{The objective results for infrared and visible image fusion and super resolution ($\bm{\times8}$, \textcolor{red}{Red}: the best, \textcolor{blue}{Blue}: the second, \textcolor{green}{Green}: the third).}
\label{tab:1}
\resizebox{\textwidth}{!}{%
\begin{tabular}{lllllll|lllllll}
\hline
\multirow{2}{*}{Methods} &
  \multicolumn{6}{c|}{Dataset: MSRS fusion dataset (\textbf{VI and IR blurred scale:8})} &
  \multirow{2}{*}{Methods} &
  \multicolumn{6}{c}{Dataset: MSRS fusion dataset (\textbf{VI blurred scale: 8})} \\
 &
  VIF↑ &
  Q\textsuperscript{AB/F}↑ &
  SSIM↑ &
  PSNR↑ &
  LPIPS↓ &
  MSE↓ &
   &
  VIF↑ &
  Q\textsuperscript{AB/F}↑ &
  SSIM↑ &
  PSNR↑ &
  LPIPS↓ &
  MSE↓ \\ \hline
CDD\textcolor{top4}{+SR3} &
  \textcolor{top2}{0.465} &
  \textcolor{top2}{0.417} &
  1.139 &
  15.944 &
  0.272 &
  2555.251 &
  CDD\textcolor{top4}{+SR3} &
  0.465 &
  0.417 &
  1.139 &
  15.944 &
  0.272 &
  2555.251 \\
Reconet\textcolor{top4}{+SR3} &
  0.335 &
  0.315 &
  0.561 &
  15.221 &
  0.320 &
  2040.049 &
  Reconet\textcolor{top4}{+SR3} &
  0.347 &
  0.323 &
  0.577 &
  15.325 &
  0.318 &
  2024.445 \\
IGNet\textcolor{top4}{+SR3} &
  0.364 &
  0.319 &
  0.885 &
  \textcolor{top2}{16.644} &
  0.281 &
  \textcolor{top3}{2000.401} &
  IGNet\textcolor{top4}{+SR3} &
  0.485 &
  0.403 &
  0.978 &
  \textcolor{top2}{16.889} &
  0.264 &
  \textcolor{top3}{1948.556} \\
UMF\textcolor{top4}{+SR3} &
  0.241 &
  0.219 &
  0.961 &
  15.742 &
  \textcolor{top1}{0.257} &
  \textcolor{top1}{1614.521} &
  UMF\textcolor{top4}{+SR3} &
  0.311 &
  0.264 &
  1.026 &
  15.902 &
  \textcolor{top3}{0.247} &
  \textcolor{top1}{1582.429} \\
LRRNet\textcolor{top4}{+SR3} &
  0.359 &
  0.367 &
  0.985 &
  15.607 &
  0.266 &
  2399.006 &
  LRRNet\textcolor{top4}{+SR3} &
  0.363 &
  0.380 &
  0.979 &
  15.775 &
  0.260 &
  2350.511 \\
Tardal\textcolor{top4}{+SR3} &
  0.129 &
  0.133 &
  0.191 &
  14.008 &
  0.337 &
  3199.039 &
  Tardal\textcolor{top4}{+SR3} &
  0.174 &
  0.166 &
  0.231 &
  14.210 &
  0.325 &
  3133.861 \\
TGFuse\textcolor{top4}{+SR3} &
  0.454 &
  0.414 &
  1.146 &
  16.233 &
  0.273 &
  2454.667 &
  TGFuse\textcolor{top4}{+SR3} &
  0.527 &
  \textcolor{top3}{0.483} &
  1.207 &
  16.372 &
  0.264 &
  2428.202 \\
DDFM\textcolor{top4}{+SR3} &
  0.397 &
  0.356 &
  1.136 &
  \textcolor{top1}{17.407} &
  \textcolor{top2}{0.258} &
  \textcolor{top2}{1770.990} &
  DDFM\textcolor{top4}{+SR3} &
  0.493 &
  0.418 &
  \textcolor{top2}{1.226} &
  \textcolor{top1}{17.829} &
  \textcolor{top2}{0.246} &
  \textcolor{top2}{1690.696} \\
DIF\textcolor{top4}{+SR3} &
  0.462 &
  0.414 &
  \textcolor{top3}{1.157} &
  15.887 &
  \textcolor{top2}{0.258} &
  2620.435 &
  DIF\textcolor{top4}{+SR3} &
  \textcolor{top3}{0.539} &
  \textcolor{top3}{0.483} &
  1.214 &
  16.013 &
  0.250 &
  2593.641 \\
CoCoNet\textcolor{top4}{+SR3} &
   0.361&
   0.267&
   0.612& 
   9.612& 
   0.384& 
   7245.056 &
   CoCoNet\textcolor{top4}{+SR3} &
   0.421 &
   0.290 &
   0.658 &
   9.651 &
   0.376 &
   7186.677\\
PromptFusion\textcolor{top4}{+SR3} &
  \textcolor{top2}{0.465} &
  \textcolor{top3}{0.415} &
  \textcolor{top2}{1.163} &
  \textcolor{top3}{16.238} &
  0.267 &
  2470.622 &
  PromptFusion\textcolor{top4}{+SR3} &
  \textcolor{top2}{0.552} &
\textcolor{top2}{0.495} &
 \textcolor{top3}{1.224} &
 \textcolor{top3}{16.384} &
0.260 &
2436.603 \\
MetaFusion &
  0.102 &
  0.055 &
  1.014 &
  13.667 &
  0.457 &
  2098.653 &
 MetaFusion &
   0.368&
   0.158&
   1.119&
   14.160&
   0.332&
   2063.281\\
Our FS-Diff &
  \textcolor{top1}{0.486}&
  \textcolor{top1}{0.481} &
  \textcolor{top1}{1.165} &
  15.858 &
  0.262 &
  2632.360 &
  Our FS-Diff &
  \textcolor{top1}{0.759} &
  \textcolor{top1}{0.673} &
  \textcolor{top1}{1.277} &
  16.123 &
  \textcolor{top1}{0.231} &
  2533.818 \\ \hline
\multirow{2}{*}{Methods} &
  \multicolumn{6}{c|}{Dataset: M\textsuperscript{3}FD fusion dataset (\textbf{VI blurred scale: 8})} &
  \multirow{2}{*}{Methods} &
  \multicolumn{6}{c}{Dataset: M\textsuperscript{3}FD fusion dataset (\textbf{VI and IR blurred scale: 8})} \\
 &
  VIF↑ &
  Q\textsuperscript{AB/F}↑ &
  SSIM↑ &
  PSNR↑ &
  LPIPS↓ &
  MSE↓ &
   &
  VIF↑ &
  Q\textsuperscript{AB/F}↑ &
  SSIM↑ &
  PSNR↑ &
  LPIPS↓ &
  MSE↓ \\ \hline
CDD\textcolor{top4}{+SR3} &
  0.303 &
  0.363 &
  0.946 &
  \textcolor{top3}{13.173} &
  0.296 &
  3844.906 &
  CDD\textcolor{top4}{+SR3} &
  0.251 &
  0.301 &
  0.892 &
  13.141 &
  0.303 &
  3868.727 \\
Reconet\textcolor{top4}{+SR3} &
  0.251 &
  0.293 &
  0.968 &
  12.037 &
  0.337 &
  3157.553 &
  Reconet\textcolor{top4}{+SR3} &
  0.226 &
  0.271 &
  0.935 &
  12.024 &
  0.339 &
  3173.556 \\
IGNet\textcolor{top4}{+SR3} &
  \textcolor{top2}{0.463} &
  \textcolor{top2}{0.436} &
  0.996 &
  12.146 &
  0.261 &
  4283.597 &
  IGNet\textcolor{top4}{+SR3} &
  \textcolor{top3}{0.334} &
  \textcolor{top3}{0.321} &
  0.908 &
  12.084 &
  \textcolor{top3}{0.281} &
  4362.066 \\
UMF\textcolor{top4}{+SR3} &
  \textcolor{top3}{0.442} &
  0.388 &
  \textcolor{top2}{1.135} &
  \textcolor{top2}{13.617} &
  \textcolor{top2}{0.233} &
  \textcolor{top1}{2390.301} &
  UMF\textcolor{top4}{+SR3} &
  \textcolor{top1}{0.341} &
  0.302 &
  \textcolor{top1}{1.067} &
  \textcolor{top2}{13.670} &
  \textcolor{top1}{0.246} &
  \textcolor{top1}{2427.097} \\
LRRNet\textcolor{top4}{+SR3} &
  0.187 &
  0.266 &
  0.874 &
  13.161 &
  0.299 &
  3475.444 &
  LRRNet\textcolor{top4}{+SR3} &
  0.175 &
  0.250 &
  0.842 &
  13.028 &
  0.306 &
  3576.946 \\
Tardal\textcolor{top4}{+SR3} &
  0.272 &
  0.272 &
  0.494 &
  10.108 &
  0.435 &
  6311.403 &
  Tardal\textcolor{top4}{+SR3} &
  0.192 &
  0.233 &
  0.420 &
  10.016 &
  0.448 &
  6472.144 \\
TGFuse\textcolor{top4}{+SR3} &
  0.279 &
  \textcolor{top3}{0.394} &
  0.965 &
  13.017 &
  0.279 &
  3951.056 &
  TGFuse\textcolor{top4}{+SR3} &
  0.232 &
  \textcolor{top2}{0.326} &
  0.922 &
  13.006 &
  0.288 &
  3979.542 \\
DDFM\textcolor{top4}{+SR3} &
  0.425 &
  0.391 &
  1.041 &
  \textcolor{top1}{14.162} &
  \textcolor{top3}{0.256} &
  \textcolor{top2}{2820.482} &
  DDFM\textcolor{top4}{+SR3} &
  \textcolor{top2}{0.340} &
  0.307 &
  \textcolor{top3}{0.974} &
  \textcolor{top1}{14.117} &
  \textcolor{top2}{0.268} &
  \textcolor{top2}{2841.389} \\
DIF\textcolor{top4}{+SR3} &
  0.298 &
  0.384 &
  0.977 &
  12.469 &
  0.297 &
  4372.365 &
  DIF\textcolor{top4}{+SR3} &
  0.252 &
  0.320 &
  0.932 &
  12.441 &
  0.306 &
  4406.081 \\
CoCoNet\textcolor{top4}{+SR3} &
  0.335 &
  0.284 &
  0.704 &
  11.690 &
  0.374 &
  4463.171 &
  CoCoNet\textcolor{top4}{+SR3} &
  0.298 &
  0.245 &
  0.658 &
  11.660 &
  0.381 &
  4497.379\\
PromptFusion\textcolor{top4}{+SR3} &
0.295  &
0.347  &
0.936  &
13.143  &
0.299  &
3802.862  &
PromptFusion\textcolor{top4}{+SR3} &
0.245  &
0.289  &
0.889  &
\textcolor{top3}{13.160}  &
0.305  &
3788.804  \\
MetaFusion &
   0.362&
   0.350&
   \textcolor{top3}{1.084}&
   11.939&
   0.312&
   \textcolor{top3}{3092.041}&
 MetaFusion &
   0.191&
   0.186&
   0.945&
   11.169&
   0.491&
   \textcolor{top3}{3159.876}\\
Our FS-Diff &
  \textcolor{top1}{0.563} &
  \textcolor{top1}{0.531} &
  \textcolor{top1}{1.271} &
  12.828 &
  \textcolor{top1}{0.212} &
  4044.098 &
  Our FS-Diff &
  0.331 &
  \textcolor{top1}{0.364} &
  \textcolor{top2}{1.025} &
  12.723 &
  0.287 &
  4278.198 \\ \hline
\multirow{2}{*}{Methods} &
  \multicolumn{6}{c|}{Dataset: AVMS fusion dataset (\textbf{VI blurred scale: 8})} &
  \multirow{2}{*}{Methods} &
  \multicolumn{6}{c}{Dataset: AVMS fusion dataset (\textbf{IR blurred scale:8})} \\
 &
  VIF↑ &
  Q\textsuperscript{AB/F}↑ &
  SSIM↑ &
  PSNR↑ &
  LPIPS↓ &
  MSE↓ &
   &
  VIF↑ &
  Q\textsuperscript{AB/F}↑ &
  SSIM↑ &
  PSNR↑ &
  LPIPS↓ &
  MSE↓ \\ \hline
CDD\textcolor{top4}{+SR3} &
  \textcolor{top3}{0.518} &
  0.493 &
  1.303 &
  15.173 &
  0.236 &
  2414.617 &
  CDD\textcolor{top4}{+SR3} &
  \textcolor{top3}{0.590} &
  0.517 &
  1.306 &
  15.167 &
  0.231 &
  2408.968 \\
Reconet\textcolor{top4}{+SR3} &
  0.446 &
  0.424 &
  1.275 &
  14.804 &
  0.322 &
  \textcolor{top2}{1818.363} &
  Reconet\textcolor{top4}{+SR3} &
  0.452 &
  0.431 &
  1.281 &
  14.828 &
  0.315 &
  \textcolor{top2}{1816.276} \\
IGNet\textcolor{top4}{+SR3} &
  0.502 &
  0.462 &
  1.096 &
  13.171 &
  0.249 &
  3183.666 &
  IGNet\textcolor{top4}{+SR3} &
  0.462 &
  0.446 &
  1.096 &
  13.189 &
  0.266 &
  3182.914 \\
UMF\textcolor{top4}{+SR3} &
  0.507 &
  0.457 &
  \textcolor{top1}{1.377} &
  \textcolor{top2}{15.809} &
  \textcolor{top2}{0.211} &
  \textcolor{top1}{1389.422} &
  UMF\textcolor{top4}{+SR3} &
  0.493 &
  0.455 &
  \textcolor{top1}{1.386} &
  \textcolor{top2}{15.838} &
  \textcolor{top2}{0.214} &
  \textcolor{top1}{1386.266} \\
LRRNet\textcolor{top4}{+SR3} &
  0.408 &
  0.400 &
  1.140 &
  14.938 &
  0.252 &
  2536.516 &
  LRRNet\textcolor{top4}{+SR3} &
  0.461 &
  0.428 &
  1.195 &
  15.053 &
  0.244 &
  2459.409 \\
Tardal\textcolor{top4}{+SR3} &
  0.341 &
  0.267 &
  0.635 &
  10.825 &
  0.404 &
  5132.788 &
  Tardal\textcolor{top4}{+SR3} &
  0.322 &
  0.276 &
  0.648 &
  10.800 &
  0.399 &
  5059.341 \\
TGFuse\textcolor{top4}{+SR3} &
  0.507 &
  \textcolor{top1}{0.546} &
  1.300 &
  15.098 &
  0.230 &
  2281.759 &
  TGFuse\textcolor{top4}{+SR3} &
  \textcolor{top2}{0.593} &
  \textcolor{top1}{0.574} &
  1.312 &
  15.085 &
  \textcolor{top3}{0.220} &
  2299.459 \\
DDFM\textcolor{top4}{+SR3} &
  0.498 &
  0.481 &
  \textcolor{top2}{1.337} &
  \textcolor{top1}{16.030} &
  \textcolor{top3}{0.224} &
  1906.830 &
  DDFM\textcolor{top4}{+SR3} &
  0.501 &
  0.487 &
  \textcolor{top2}{1.344} &
  \textcolor{top1}{16.013} &
  0.223 &
  1918.311 \\
DIF\textcolor{top4}{+SR3} &
  0.484 &
  \textcolor{top3}{0.516} &
  1.270 &
  14.629 &
  0.252 &
  2538.968 &
  DIF\textcolor{top4}{+SR3} &
  0.501 &
  0.527 &
  1.278 &
  14.647 &
  0.241 &
  2546.911 \\  
CoCoNet\textcolor{top4}{+SR3}&
  0.437 &
  0.344 &
  0.998 &
  13.069 &
  0.304 &
  3216.136 &
  CoCoNet\textcolor{top4}{+SR3}&
  0.465 &
  0.350 &
  1.004 &
  13.070 &
  0.297 &
  3222.768  \\ 
PromptFusion\textcolor{top4}{+SR3}&
\textcolor{top2}{0.551} &
0.509 &
1.307 &
\textcolor{top3}{15.281} &
0.232 &
2480.833 &
PromptFusion\textcolor{top4}{+SR3}&
\textcolor{top1}{0.641} &
\textcolor{top2}{0.540} &
1.313 &
\textcolor{top3}{15.271} &
0.226 &
2462.413 \\
MetaFusion &
   0.427&
   0.359&
   1.108&
   13.572&
   0.307&
   \textcolor{top3}{1897.778}&
 MetaFusion &
   0.433&
   0.363&
   1.127&
   13.959&
   0.293&
   \textcolor{top3}{1886.390}\\
Our FS-Diff &
  \textcolor{top1}{0.559} &
  \textcolor{top2}{0.534} &
  \textcolor{top3}{1.334} &
  14.867 &
  \textcolor{top1}{0.209} &
  2389.134 &
  Our FS-Diff &
  0.552 &
  \textcolor{top3}{0.532} &
  \textcolor{top2}{1.344} &
  14.756 &
  \textcolor{top1}{0.211} &
  2368.634 \\ \hline
\multirow{2}{*}{Methods} &
  \multicolumn{6}{c|}{Dataset: LLVIP fusion dataset (\textbf{IR blurred scale: 8})} &
  \multirow{2}{*}{Methods} &
  \multicolumn{6}{c}{Dataset: LLVIP fusion dataset (\textbf{VI blurred scale: 8})} \\
 &
  VIF↑ &
  Q\textsuperscript{AB/F}↑ &
  SSIM↑ &
  PSNR↑ &
  LPIPS↓ &
  MSE↓ &
   &
  VIF↑ &
  Q\textsuperscript{AB/F}↑ &
  SSIM↑ &
  PSNR↑ &
  LPIPS↓ &
  MSE↓ \\ \hline
CDD\textcolor{top4}{+SR3} &
  \textcolor{top1}{0.571} &
  \textcolor{top2}{0.546} &
  \textcolor{top2}{1.148} &
  14.366 &
  \textcolor{top3}{0.277} &
  2511.818 &
  CDD\textcolor{top4}{+SR3} &
  \textcolor{top1}{0.666} &
  \textcolor{top1}{0.630} &
  \textcolor{top2}{1.190} &
  14.342 &
  \textcolor{top3}{0.267} &
  2538.224 \\
Reconet\textcolor{top4}{+SR3} &
  0.363 &
  0.296 &
  0.751 &
  12.440 &
  0.360 &
  2284.514 &
  Reconet\textcolor{top4}{+SR3} &
  0.369 &
  0.302 &
  0.760 &
  12.449 &
  0.359 &
  2258.865 \\
IGNet\textcolor{top4}{+SR3} &
  0.481 &
  0.480 &
  1.047 &
  \textcolor{top2}{14.929} &
  0.284 &
  2114.455 &
  IGNet\textcolor{top4}{+SR3} &
  \textcolor{top3}{0.557} &
  0.562 &
  1.095 &
  \textcolor{top2}{15.008} &
  0.272 &
  2090.975 \\
UMF\textcolor{top4}{+SR3} &
  0.364 &
  0.331 &
  1.057 &
  \textcolor{top3}{14.926} &
  0.283 &
  \textcolor{top1}{1498.730} &
  UMF\textcolor{top4}{+SR3} &
  0.390 &
  0.368 &
  1.080 &
  14.923 &
  0.276 &
  \textcolor{top1}{1460.715} \\
LRRNet\textcolor{top4}{+SR3} &
  0.359 &
  0.341 &
  0.929 &
  \textcolor{top1}{15.573} &
  0.297 &
  \textcolor{top2}{1838.926} &
  LRRNet\textcolor{top4}{+SR3} &
  0.348 &
  0.336 &
  0.932 &
  \textcolor{top1}{15.576} &
  0.295 &
  \textcolor{top2}{1843.225} \\
Tardal\textcolor{top4}{+SR3} &
  0.223 &
  0.262 &
  0.465 &
  12.933 &
  0.421 &
  3225.795 &
  Tardal\textcolor{top4}{+SR3} &
  0.252 &
  0.294 &
  0.497 &
  13.072 &
  0.413 &
  3117.933 \\
TGFuse\textcolor{top4}{+SR3} &
  \textcolor{top3}{0.534} &
  \textcolor{top1}{0.549} &
  1.118 &
  14.781 &
  0.283 &
  \textcolor{top3}{2077.500} &
  TGFuse\textcolor{top4}{+SR3} &
  0.555 &
  \textcolor{top3}{0.597} &
  1.152 &
  14.834 &
  0.273 &
  \textcolor{top3}{2028.494} \\
DDFM\textcolor{top4}{+SR3} &
  0.508 &
  0.485 &
  1.135 &
  14.877 &
  0.276 &
  2437.723 &
  DDFM\textcolor{top4}{+SR3} &
  \textcolor{top3}{0.557} &
  0.541 &
  \textcolor{top3}{1.173} &
  \textcolor{top3}{14.952} &
  0.268 &
  2423.511 \\
DIF\textcolor{top4}{+SR3} &
  0.507 &
  0.511 &
  1.100 &
  13.792 &
  0.299 &
  2579.120 &
  DIF\textcolor{top4}{+SR3} &
  0.546 &
  0.562 &
  1.134 &
  13.835 &
  0.293 &
  2538.445 \\
CoCoNet\textcolor{top4}{+SR3}&
  0.465 &
  0.388 &
  0.884 &
  11.088 &
  0.325 &
  5370.172 &
  CoCoNet\textcolor{top4}{+SR3}&
  0.474 &
  0.419 &
  0.914 &
  11.127 &
  0.320 &
  5318.431 \\
PromptFusion\textcolor{top4}{+SR3}&
\textcolor{top2}{0.567} &
\textcolor{top3}{0.543} &
\textcolor{top1}{1.169} &
14.899 &
\textcolor{top2}{0.271} &
2256.018 &
PromptFusion\textcolor{top4}{+SR3}&
\textcolor{top2}{0.651} &
\textcolor{top2}{0.622} &
\textcolor{top1}{1.208} &
14.905 &
\textcolor{top2}{0.262} &
2272.659 \\
MetaFusion &
   0.385&
   0.221&
   0.999&
   12.185&
   0.382&
   2217.690&
 MetaFusion &
   0.242&
   0.194&
   0.921&
   12.476&
   0.385&
   2100.118\\
Our FS-Diff &
  0.480 &
  0.481 &
  \textcolor{top3}{1.140} &
  13.806 &
  \textcolor{top1}{0.262} &
  2488.696 &
  Our FS-Diff &
  0.473 &
  0.481 &
  1.133 &
  13.872 &
  \textcolor{top1}{0.259} &
  2511.560 \\ \hline
\end{tabular}%
}
\end{table*}


\begin{table}[]
\centering
\caption{The objective results for infrared and visible image fusion and super resolution ($\bm{\times2}$, $\bm{\times4}$, \textcolor{red}{Red}: the best, \textcolor{blue}{Blue}: the second, \textcolor{green}{Green}: the third).}
\label{tab:add}
\resizebox{\textwidth}{!}{%
%\begin{table}[]
\begin{tabular}{lllllll|lllllll}
\hline
\multirow{2}{*}{Methods} &
  \multicolumn{6}{c|}{Dataset: MSRS fusion datase(\textbf{VI blurred scale:2})} &
  \multirow{2}{*}{Methods} &
  \multicolumn{6}{c}{Dataset:  M\textsuperscript{3}FD fusion datase(\textbf{VI and IR blurred scale:2})} \\ 
 &
  VIF↑ &
  QAB/F↑ &
  SSIM↑ &
  PSNR↑ &
  LPIPS↓ &
  MSE↓ &
   &
  VIF↑ &
  QAB/F↑ &
  SSIM↑ &
  PSNR↑ &
  LPIPS↓ &
  MSE↓ \\ \hline
CDD\textcolor{top4}{+SR3} &
  0.322 &
  0.266 &
  1.014 &
  15.283 &
  0.288 &
  3626.851 &
  CDD\textcolor{top4}{+SR3} &
  0.391 &
  0.349 &
  0.992 &
  10.860 &
  0.270 &
  6478.874 \\
Reconet\textcolor{top4}{+SR3} &
  0.332 &
  0.275 &
  0.687 &
  15.393 &
  0.320 &
  2102.230 &
  Reconet\textcolor{top4}{+SR3} &
  0.335 &
  0.339 &
   \textcolor{top3}{ 1.128} &
  12.227 &
  0.298 &
   \textcolor{top3}{ 3112.319} \\
IGNet\textcolor{top4}{+SR3} &
  0.258 &
  0.196 &
  0.719 &
  14.744 &
  0.381 &
  2289.869 &
  IGNet\textcolor{top4}{+SR3} &
  0.199 &
  0.278 &
  0.645 &
  11.227 &
  0.425 &
  4792.991 \\
UMF\textcolor{top4}{+SR3} &
  0.284 &
  0.188 &
  1.057 &
   \textcolor{top2}{ 16.276} &
   \textcolor{top2}{ 0.246} &
  %{\color[HTML]{FE0000} 1542.097} &
  \textcolor{top1}{1542.097}&
  UMF\textcolor{top4}{+SR3} &
  0.359 &
  0.302 &
   \textcolor{top2}{ 1.178} &
   \textcolor{top3}{ 13.405} &
   \textcolor{top1}{ 0.222} &
   \textcolor{top1}{ 2420.237} \\
LRRNet\textcolor{top4}{+SR3} &
  0.286 &
  0.280 &
  0.901 &
  15.419 &
  0.270 &
  2460.521 &
  LRRNet\textcolor{top4}{+SR3} &
  0.288 &
  0.318 &
  0.995 &
  12.988 &
  0.252 &
  3609.997 \\
Tardal\textcolor{top4}{+SR3} &
  0.391 &
  0.256 &
  0.864 &
  12.444 &
  0.286 &
  3779.646 &
  Tardal\textcolor{top4}{+SR3} &
   \textcolor{top2}{ 0.417} &
  0.323 &
  1.093 &
  12.861 &
  0.288 &
  3783.758 \\
TGFuse\textcolor{top4}{+SR3} &
  0.401 &
   \textcolor{top3}{ 0.359} &
  1.019 &
  14.803 &
  0.275 &
  2925.301 &
  TGFuse\textcolor{top4}{+SR3} &
  0.318 &
   \textcolor{top2}{ 0.379} &
  1.087 &
  12.755 &
   \textcolor{top3}{ 0.235} &
  4138.130 \\
DDFM\textcolor{top4}{+SR3} &
   \textcolor{top3}{ 0.410} &
  0.313 &
  1.013 &
   \textcolor{top1}{ 16.790} &
  0.294 &
   \textcolor{top2}{ 1901.532} &
  DDFM\textcolor{top4}{+SR3} &
  0.372 &
  0.329 &
  1.090 &
   \textcolor{top1}{ 14.143} &
  0.242 &
   \textcolor{top2}{ 2869.899} \\
DIF\textcolor{top4}{+SR3} &
   \textcolor{top2}{ 0.428} &
   \textcolor{top2}{ 0.364} &
   1.058 &
  14.578 &
   \textcolor{top3}{ 0.248} &
  3091.676 &
  DIF\textcolor{top4}{+SR3} &
  0.350 &
   \textcolor{top3}{ 0.374} &
  1.102 &
  12.151 &
  0.246 &
  4600.052 \\
CoCoNet\textcolor{top4}{+SR3}&
  0.331 &
  0.228 &
  0.555 &
  9.528 &
  0.413 &
  7406.531 &
  CoCoNet\textcolor{top4}{+SR3}&
  \textcolor{top3}{0.392} &
  0.282 &
  0.797 &
  11.839 &
  0.324 &
  4307.486 \\
PromptFuison\textcolor{top4}{+SR3}&
0.394 &
0.350 &
\textcolor{top3}{1.070} &
15.488 &
0.275 &
2697.993 &
PromptFuison\textcolor{top4}{+SR3}&
0.380 &
0.354 &
1.052 &
12.914 &
0.248 &
4016.798 \\
MetaFusion &
  0.287 &
  0.179 &
   \textcolor{top2}{ 1.080} &
  15.232 &
  0.302 &
   \textcolor{top3}{ 1990.558} &
  MetaFusion &
  0.347 &
  0.328 &
  1.089 &
  11.661 &
  0.277 &
  3225.636 \\
Our FS-Diff &
   \textcolor{top1}{ 0.834} &
   \textcolor{top1}{ 0.693} &
   \textcolor{top1}{ 1.274} &
   \textcolor{top3}{ 16.085} &
   \textcolor{top1}{ 0.231} &
  2500.553 &
  Our FS-Diff &
   \textcolor{top1}{ 0.606} &
   \textcolor{top1}{ 0.541} &
   \textcolor{top1}{ 1.226} &
   \textcolor{top2}{ 13.447} &
   \textcolor{top2}{ 0.224} &
  3562.457 \\ \hline
\multirow{2}{*}{Methods} &
  \multicolumn{6}{c|}{Dataset: LLVIP fusion datase(\textbf{IR blurred scale:2})} &
  \multirow{2}{*}{Methods} &
  \multicolumn{6}{c}{Dataset: AVMS fusion datase(\textbf{VI and IR blurred scale:2})} \\
 &
  VIF↑ &
  QAB/F↑ &
  SSIM↑ &
  PSNR↑ &
  LPIPS↓ &
  MSE↓ &
   &
  VIF↑ &
  QAB/F↑ &
  SSIM↑ &
  PSNR↑ &
  LPIPS↓ &
  MSE↓ \\ \hline
CDD\textcolor{top4}{+SR3} &
   \textcolor{top1}{ 0.650} &
   0.593 &
   1.156 &
  13.487 &
  0.277 &
  3075.494 &
  CDD\textcolor{top4}{+SR3} &
   \textcolor{top2}{ 0.482} &
  0.410 &
  1.134 &
  12.822 &
  0.263 &
  3866.371 \\
Reconet\textcolor{top4}{+SR3} &
  0.365 &
  0.297 &
  0.727 &
  12.210 &
  0.364 &
  2386.032 &
  Reconet\textcolor{top4}{+SR3} &
  0.433 &
  0.400 &
   \textcolor{top3}{ 1.246} &
  14.684 &
  0.320 &
   \textcolor{top2}{ 1902.222} \\
IGNet\textcolor{top4}{+SR3} &
  0.281 &
  0.322 &
  0.784 &
  13.478 &
  0.371 &
  2570.188 &
  IGNet\textcolor{top4}{+SR3} &
  0.278 &
  0.308 &
  0.785 &
  12.015 &
  0.390 &
  3687.439 \\
UMF\textcolor{top4}{+SR3} &
  0.377 &
  0.340 &
  1.073 &
  14.816 &
  0.279 &
   \textcolor{top1}{ 1446.383} &
  UMF\textcolor{top4}{+SR3} &
  0.431 &
  0.382 &
   \textcolor{top2}{ 1.286} &
   \textcolor{top2}{ 15.404} &
   \textcolor{top1}{ 0.220} &
   \textcolor{top1}{ 1475.811} \\
LRRNet\textcolor{top4}{+SR3} &
  0.347 &
  0.337 &
  0.905 &
   \textcolor{top1}{ 15.465} &
  0.306 &
   \textcolor{top3}{ 1893.073} &
  LRRNet\textcolor{top4}{+SR3} &
  0.378 &
  0.363 &
  1.092 &
  14.920 &
  0.252 &
  2501.392 \\
Tardal\textcolor{top4}{+SR3} &
  0.447 &
  0.354 &
  1.062 &
  13.489 &
  0.320 &
  2851.051 &
  Tardal\textcolor{top4}{+SR3} &
  0.435 &
  0.378 &
  1.197 &
  14.190 &
  0.299 &
  2818.125 \\
TGFuse\textcolor{top4}{+SR3} &
   0.576 &
   \textcolor{top2}{ 0.597} &
   \textcolor{top3}{ 1.159} &
   14.957 &
   \textcolor{top3}{ 0.272} &
  1947.154 &
  TGFuse\textcolor{top4}{+SR3} &
  0.440 &
   \textcolor{top2}{ 0.451} &
  1.207 &
  14.676 &
  0.232 &
  2482.844 \\
DDFM\textcolor{top4}{+SR3} &
  0.484 &
  0.488 &
  1.125 &
   \textcolor{top2}{ 15.345} &
  0.295 &
  2152.899 &
  DDFM\textcolor{top4}{+SR3} &
  0.423 &
  0.410 &
  1.240 &
   \textcolor{top1}{ 15.933} &
   \textcolor{top3}{ 0.232} &
   \textcolor{top3}{ 1933.640} \\
DIF\textcolor{top4}{+SR3} &
  0.534 &
  0.560 &
  1.115 &
  13.586 &
  0.295 &
  2662.501 &
  DIF\textcolor{top4}{+SR3} &
   0.450 &
   \textcolor{top3}{ 0.447} &
  1.217 &
  14.278 &
  0.249 &
  2751.751 \\
CoCoNet\textcolor{top4}{+SR3}&
  0.484 &
  0.413 &
  0.912 &
  11.123 &
  0.320 &
  5329.684 &
  CoCoNet\textcolor{top4}{+SR3}&
  0.388 &
  0.306 &
  0.917 &
  12.974 &
  0.310 &
  3288.426 \\
PromptFusion\textcolor{top4}{+SR3}&
\textcolor{top2}{0.617} &
\textcolor{top3}{0.595} &
\textcolor{top1}{1.209} &
\textcolor{top3}{15.201} &
\textcolor{top2}{0.263} &
2125.803 &
PromptFusion\textcolor{top4}{+SR3}&
\textcolor{top3}{0.481} &
0.417 &
1.180 &
14.921 &
0.242 &
2611.946 \\
MetaFusion &
  0.351 &
  0.295 &
  1.015 &
  13.601 &
  0.341 &
   \textcolor{top2}{ 1782.251} &
  MetaFusion &
  0.424 &
  0.383 &
  1.201 &
  13.862 &
  0.269 &
  1975.047 \\
Our FS-Diff &
   \textcolor{top3}{ 0.607} &
   \textcolor{top1}{ 0.621} &
   \textcolor{top2}{ 1.182} &
  13.953 &
   \textcolor{top1}{ 0.258} &
  2704.236 &
  Our FS-Diff &
   \textcolor{top1}{ 0.550} &
   \textcolor{top1}{ 0.510} &
   \textcolor{top1}{ 1.307} &
   \textcolor{top3}{ 15.163} &
   \textcolor{top2}{ 0.222} &
  2390.913 \\ \hline
\multirow{2}{*}{Methods} &
  \multicolumn{6}{c|}{Dataset: MSRS fusion datase(\textbf{VI blurred scale:4})} &
  \multirow{2}{*}{Methods} &
  \multicolumn{6}{c}{Dataset:  M\textsuperscript{3}FD fusion datase(\textbf{VI and IR blurred scale:4})} \\
 &
  VIF↑ &
  QAB/F↑ &
  SSIM↑ &
  PSNR↑ &
  LPIPS↓ &
  MSE↓ &
   &
  VIF↑ &
  QAB/F↑ &
  SSIM↑ &
  PSNR↑ &
  LPIPS↓ &
  MSE↓ \\ \hline
CDD\textcolor{top4}{+SR3} &
  0.347 &
  0.278 &
  0.985 &
  15.512 &
  0.293 &
  3344.432 &
  CDD\textcolor{top4}{+SR3} &
   \textcolor{top3}{ 0.452} &
  0.374 &
  1.022 &
  11.032 &
  0.267 &
  6187.481 \\
Reconet\textcolor{top4}{+SR3} &
  0.323 &
  0.268 &
  0.558 &
  15.262 &
  0.321 &
  2039.084 &
  Reconet\textcolor{top4}{+SR3} &
  0.360 &
  0.352 &
  1.121 &
  12.126 &
  0.293 &
   \textcolor{top3}{ 3053.902} \\
IGNet\textcolor{top4}{+SR3} &
  0.272 &
  0.210 &
  0.731 &
  14.909 &
  0.374 &
  2234.577 &
  IGNet\textcolor{top4}{+SR3} &
  0.215 &
  0.292 &
  0.660 &
  11.288 &
  0.416 &
  4795.159 \\
UMF\textcolor{top4}{+SR3} &
  0.294 &
  0.227 &
  1.001 &
  15.882 &
   \textcolor{top2}{ 0.252} &
   \textcolor{top1}{ 1594.493} &
  UMF\textcolor{top4}{+SR3} &
  0.405 &
  0.335 &
   \textcolor{top2}{ 1.206} &
   \textcolor{top2}{ 13.473} &
   \textcolor{top1}{ 0.221} &
   \textcolor{top1}{ 2429.874} \\
LRRNet\textcolor{top4}{+SR3} &
  0.319 &
  0.303 &
  0.930 &
  15.723 &
  0.266 &
  2381.735 &
  LRRNet\textcolor{top4}{+SR3} &
  0.330 &
  0.339 &
  1.048 &
  13.108 &
  0.247 &
  3535.829 \\
Tardal\textcolor{top4}{+SR3} &
  0.418 &
  0.264 &
  0.873 &
  12.538 &
  0.282 &
  3733.281 &
  Tardal\textcolor{top4}{+SR3} &
   \textcolor{top2}{ 0.453} &
  0.344 &
  1.118 &
  12.951 &
  0.284 &
  3745.289 \\
TGFuse\textcolor{top4}{+SR3} &
   0.475 &
   0.396 &
   1.154 &
   \textcolor{top2}{ 16.255} &
  0.270 &
  2477.432 &
  TGFuse\textcolor{top4}{+SR3} &
  0.378 &
   \textcolor{top2}{ 0.419} &
  1.131 &
  13.187 &
   \textcolor{top3}{ 0.234} &
  3832.382 \\
DDFM\textcolor{top4}{+SR3} &
  0.427 &
  0.328 &
  1.131 &
   \textcolor{top1}{ 17.772} &
  0.272 &
   \textcolor{top2}{ 1696.539} &
  DDFM\textcolor{top4}{+SR3} &
  0.414 &
  0.361 &
  1.126 &
   \textcolor{top1}{ 14.295} &
  0.241 &
   \textcolor{top2}{ 2766.588} \\
DIF\textcolor{top4}{+SR3} &
   \textcolor{top2}{ 0.500} &
   \textcolor{top3}{ 0.402} &
   \textcolor{top2}{ 1.174} &
  15.927 &
   \textcolor{top3}{ 0.253} &
  2635.370 &
  DIF\textcolor{top4}{+SR3} &
  0.399 &
   \textcolor{top3}{ 0.408} &
   \textcolor{top3}{ 1.133} &
  12.596 &
  0.247 &
  4261.666 \\
CoCoNet\textcolor{top4}{+SR3} &
  0.391 &
  0.247 &
  0.612 &
  9.622 &
  0.380 &
  7226.883 &
  CoCoNet\textcolor{top4}{+SR3} &
  0.445 &
  0.298 &
  0.856 &
  11.886 &
  0.311 &
  4262.548 \\
PromptFusion\textcolor{top4}{+SR3} &
\textcolor{top3}{0.495} &
\textcolor{top2}{0.405} &
\textcolor{top3}{1.167} &
\textcolor{top3}{16.282} &
0.267 &
2482.416 &
PromptFusion\textcolor{top4}{+SR3} &
0.444 &
0.385 &
1.108 &
13.174 &
0.242 &
3801.288 \\
MetaFusion &
  0.285 &
  0.219 &
  1.094 &
  15.434 &
  0.299 &
   \textcolor{top3}{ 1916.292} &
  MetaFusion &
  0.375 &
  0.336 &
  1.093 &
  11.783 &
  0.282 &
  3130.103 \\
Our FS-Diff &
   \textcolor{top1}{ 0.731} &
   \textcolor{top1}{ 0.669} &
   \textcolor{top1}{ 1.267} &
   16.061 &
   \textcolor{top1}{ 0.233} &
  2528.382 &
  Our FS-Diff &
   \textcolor{top1}{ 0.583} &
   \textcolor{top1}{ 0.533} &
   \textcolor{top1}{ 1.218} &
   \textcolor{top3}{ 13.440} &
   \textcolor{top2}{ 0.227} &
  3561.357 \\ \hline
\multirow{2}{*}{Methods} &
  \multicolumn{6}{c|}{Dataset: LLVIP fusion datase(\textbf{IR blurred scale:4})} &
  \multirow{2}{*}{Methods} &
  \multicolumn{6}{c}{Dataset: AVMS fusion datase(\textbf{VI and IR blurred scale:4})} \\
 &
  VIF↑ &
  QAB/F↑ &
  SSIM↑ &
  PSNR↑ &
  LPIPS↓ &
  MSE↓ &
   &
  VIF↑ &
  QAB/F↑ &
  SSIM↑ &
  PSNR↑ &
  LPIPS↓ &
  MSE↓ \\ \hline
CDD\textcolor{top4}{+SR3} &
   \textcolor{top1}{ 0.636} &
  0.561 &
  1.137 &
  13.314 &
  0.282 &
  3192.400 &
  CDD\textcolor{top4}{+SR3} &
   \textcolor{top2}{ 0.379} &
  0.304 &
  0.969 &
  12.982 &
  0.295 &
  3754.106 \\
Reconet\textcolor{top4}{+SR3} &
  0.363 &
  0.295 &
  0.731 &
  12.270 &
  0.361 &
  2359.030 &
  Reconet\textcolor{top4}{+SR3} &
  0.356 &
  0.307 &
   \textcolor{top3}{ 1.094} &
  14.372 &
  0.336 &
   \textcolor{top2}{ 1880.610} \\
IGNet\textcolor{top4}{+SR3} &
  0.277 &
  0.313 &
  0.777 &
  13.489 &
  0.370 &
  2567.333 &
  IGNet\textcolor{top4}{+SR3} &
  0.234 &
  0.250 &
  0.680 &
  11.938 &
  0.406 &
  3747.477 \\
UMF\textcolor{top4}{+SR3} &
  0.370 &
  0.334 &
  1.068 &
  14.914 &
  0.279 &
   \textcolor{top1}{ 1463.648} &
  UMF\textcolor{top4}{+SR3} &
  0.340 &
  0.279 &
   \textcolor{top2}{ 1.100} &
   \textcolor{top3}{ 14.895} &
   \textcolor{top2}{ 0.253} &
   \textcolor{top1}{ 1563.837} \\
LRRNet\textcolor{top4}{+SR3} &
  0.342 &
  0.321 &
  0.911 &
   \textcolor{top1}{ 15.534} &
  0.297 &
   \textcolor{top2}{ 1854.220} &
  LRRNet\textcolor{top4}{+SR3} &
  0.307 &
  0.283 &
  0.945 &
  14.736 &
  0.283 &
  2598.644 \\
Tardal\textcolor{top4}{+SR3} &
  0.440 &
  0.344 &
  1.053 &
  13.459 &
  0.321 &
  2892.844 &
  Tardal\textcolor{top4}{+SR3} &
  0.362 &
  0.294 &
  1.052 &
  14.111 &
  0.322 &
  2849.113 \\
TGFuse\textcolor{top4}{+SR3} &
   0.558 &
   \textcolor{top2}{ 0.571} &
   \textcolor{top3}{ 1.144} &
   14.939 &
   \textcolor{top3}{ 0.277} &
   \textcolor{top3}{ 1989.202} &
  TGFuse\textcolor{top4}{+SR3} &
  0.339 &
   \textcolor{top2}{ 0.323} &
  1.028 &
  14.754 &
   \textcolor{top3}{ 0.266} &
  2366.751 \\
DDFM\textcolor{top4}{+SR3} &
  0.475 &
  0.472 &
  1.114 &
   \textcolor{top2}{ 15.274} &
  0.295 &
  2193.437 &
  DDFM\textcolor{top4}{+SR3} &
  0.337 &
  0.304 &
  1.043 &
   \textcolor{top1}{ 15.671} &
  0.267 &
   \textcolor{top3}{ 2002.561} \\
DIF\textcolor{top4}{+SR3} &
  0.541 &
  0.554 &
 1.141 &
  14.165 &
  0.292 &
  2312.984 &
  DIF\textcolor{top4}{+SR3} &
  0.347 &
   \textcolor{top3}{ 0.320} &
  1.049 &
  14.450 &
  0.277 &
  2590.212 \\
CoCoNet\textcolor{top4}{+SR3}&
  0.476 &
  0.400 &
  0.903 &
  11.115 &
  0.322 &
  5340.444 &
  CoCoNet\textcolor{top4}{+SR3}&
  0.320& 
  0.245 &
  0.780 &
  12.796 &
  0.328 &
  3415.530 \\
PromptFusion\textcolor{top4}{+SR3} &
\textcolor{top2}{0.597} &
\textcolor{top3}{0.569} &
\textcolor{top1}{1.195} &
\textcolor{top3}{15.128} &
\textcolor{top2}{0.266} &
2148.109 &
PromptFusion\textcolor{top4}{+SR3} &
\textcolor{top3}{0.371} &
0.303 &
0.992 &
14.762 &
0.275 &
2683.096 \\
MetaFusion &
  0.334 &
  0.249 &
  1.025 &
  12.926 &
  0.347 &
  2027.974 &
  MetaFusion &
  0.340 &
  0.290 &
  1.048 &
  13.467 &
  0.291 &
  2009.857 \\
Our FS-Diff &
   \textcolor{top3}{ 0.592} &
   \textcolor{top1}{ 0.614} &
   \textcolor{top2}{ 1.179} &
  13.981 &
   \textcolor{top1}{ 0.260} &
  2681.306 &
  Our FS-Diff &
   \textcolor{top1}{ 0.534} &
   \textcolor{top1}{ 0.503} &
   \textcolor{top1}{ 1.298} &
   \textcolor{top2}{ 15.134} &
   \textcolor{top1}{ 0.229} &
  2405.701 \\ \hline

\end{tabular}%
}
\end{table}

\subsection{Medical image fusion}
\label{5.3}
% \usepackage{multirow}
% \usepackage{graphicx}

\begin{figure*}[htbp]
\centering	
\includegraphics[width=1.0\linewidth]{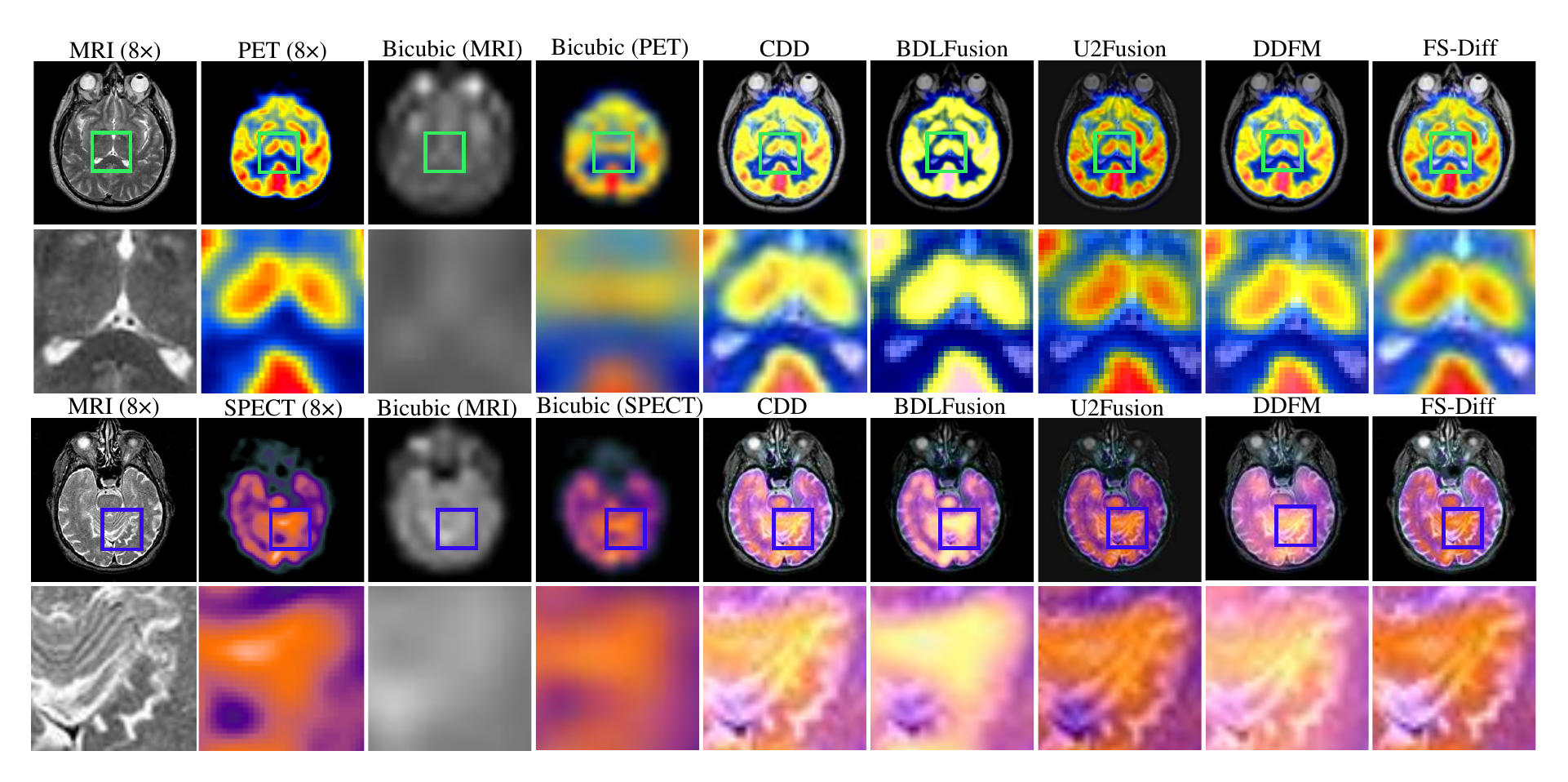}
\caption{Fusion and super-resolution (scale:8) results for FS-Diff as well as the fusion results for the compared methods on the Harvard medical image dataset.}
\label{Frame6}
\vspace{-0.5cm}
\end{figure*}

The results of the subjective and objective assessments of the MIF task, under the condition where both image modalities are blurred, are presented in Table \ref{tab:2} and Figure \ref{Frame6}. In Figure \ref{Frame6}, the comparison methods are fuzzy or even lost when processing structure and color information. For example, BDLFusion, U2Fusion, and DDPM fail to effectively retain marginal skeletal information on MRI, whereas CDD exhibits poor retention of color information and distortion on PET/SPECT images. The third column in Figure \ref{Frame6} shows the results of FS-Diff at $\bm{\times8}$ magnification (\(16^{2}\) \(\rightarrow\) \(128^{2}\)). The fusion results generated by FS-Diff retained excellent edges, contrasts, and structures.
% \begin{figure*}[t]
% \centering	
% \includegraphics[width=1.0\linewidth]{image/07.pdf}
% \caption{Effectiveness of the FAB in promoting cross-modal information fusion.}
% \label{Frame7}
% \end{figure*}

As shown in Table \ref{tab:2}, our FS-Diff consistently performed well in terms of the VIF, Q\textsuperscript{AB/F}, SSIM, PSNR, and LPIPS metrics on the MRI-PET and MRI-SPECT MIF datasets slightly lagged behind U2Fusion\rem{ on MSE. It is worth noting that}\aug{in terms of the MSE. Notably,} although the results of the comparison method were obtained via multimodal HR source image fusion and then down-sampling (\(256^2 \rightarrow 128^2\)), the subjective and objective assessments were still inferior to the results of FS-Diff (\(16^2 \rightarrow 128^2\)), which fully proved the effectiveness of FS-Diff on MIF.
%\begin{table}[H][!ht]
\begin{table}[!ht]
\centering
\caption{The objective results for medical image fusion and super resolution (\textcolor{red}{Red}: the best, \textcolor{blue}{Blue}: the second best)}
\label{tab:2}
{%
\begin{tabular}{llllllll}
\hline
\multirow{2}{*}{Methods} & \multicolumn{7}{c}{Dataset:   MRI-PET of Harvard medical image dataset}   \\
                         & Scale    & VIF↑     & Q\textsuperscript{AB/F}↑   & SSIM↑   & PSNR↑    & LPIPS↓  & MSE↓      \\ \cline{2-8} 
CDD                      & 1        & \textcolor{top2}{0.714}    & \textcolor{top2}{0.690}    & \textcolor{top2}{1.335}   & 12.889   & 0.303   & 3505.219  \\
BDLFusion                & 1        & 0.479    & 0.528    & 1.154   & 13.782   & 0.308   & 2525.134  \\
U2Fusion                 & 1        & 0.481    & 0.415    & 0.620   & 14.550   & \textcolor{top2}{0.293}   & \textcolor{top1}{1874.329}  \\
DDFM                     & 1        & 0.659    & 0.594    & 1.285   & \textcolor{top1}{14.614}   & \textcolor{top1}{0.292}   & \textcolor{top2}{2057.891}  \\
Our FS-Diff              & 8        & \textcolor{top1}{0.760}    & \textcolor{top1}{0.719}    & \textcolor{top1}{1.377}   & \textcolor{top2}{13.367}   & 0.303   & 3194.768  \\ \hline
\multirow{2}{*}{Methods} & \multicolumn{7}{c}{Dataset: MRI-SPECT of Harvard medical image   dataset} \\
                         & Scale    & VIF↑     & Q\textsuperscript{AB/F}↑   & SSIM↑   & PSNR↑    & LPIPS↓  & MSE↓      \\ \cline{2-8} 
CDD                      & 1        & \textcolor{top1}{0.709}    & \textcolor{top1}{0.642}    & \textcolor{top2}{1.468}   & 17.115   & 0.228   & 1290.771  \\
BDLFusion                & 1        & 0.474    & 0.388    & 1.361   & 16.259   & 0.231   & 1488.426  \\
U2Fusion                 & 1        & 0.412    & 0.358    & 0.613   & \textcolor{top2}{17.392}   & \textcolor{top2}{0.226}   & \textcolor{top1}{729.343}   \\
DDFM                     & 1        & 0.552    & 0.560    & 1.418   & 16.966   & 0.237   & 1289.635  \\
Our FS-Diff              & 8        & \textcolor{top2}{0.686}    & \textcolor{top2}{0.639}    & \textcolor{top1}{1.476}   & \textcolor{top1}{18.057}   & \textcolor{top1}{0.224}   & \textcolor{top2}{1022.069}  \\ \hline
\end{tabular}%
}
\end{table}

\subsection{Ablation experiments}
\label{5.4}

\subsubsection{The influence of BFM and CLSE}

\begin{table*}[]
\centering
\caption{Ablation experiment on the CLSE and BFM. (\textcolor{red}{Red}: the best)}
\label{tab:3}
\resizebox{\textwidth}{!}{%
\begin{tabular}{lllllllll|lllllllll}
\hline
 &
  \multicolumn{2}{l}{Method} &
  \multicolumn{6}{c|}{MSRS (both blurred) scale:8} &
   &
  \multicolumn{2}{l}{Method} &
  \multicolumn{6}{c}{AVMS (both blurred) scale:8} \\ \cline{2-3} \cline{11-12}
 &
  \del{CA-CLIP}\add{CLSE} &
  BFM &
  VIF↑ &
  QAB/F↑ &
  SSIM↑ &
  PSNR↑ &
  LPIPS↓ &
  MSE↓ &
   &
  \del{CA-CLIP}\add{CLSE} &
  BFM &
  VIF↑ &
  QAB/F↑ &
  SSIM↑ &
  PSNR↑ &
  LPIPS↓ &
  MSE↓ \\ \hline
1 &
  \ding{51} &
  \ding{55} &
  0.472 &
  0.479 &
  1.061 &
  13.489 &
  0.343 &
  2662.481 &
  1 &
  \ding{51} &
  \ding{55} &
  0.283 &
  0.248 &
  0.862 &
  13.817 &
  0.423 &
  2974.492 \\
2 &
  \ding{55} &
  \ding{51} &
  0.479 &
  0.471 &
  1.055 &
  13.592 &
  0.299 &
  2757.148 &
  2 &
  \ding{55} &
  \ding{51} &
  0.257 &
  0.283 &
  0.846 &
  13.574 &
  0.367 &
  2967.923 \\
3 &
  \ding{55} &
  \ding{55} &
  0.453 &
  0.461 &
  0.848 &
  12.489 &
  0.328 &
  2904.463 &
  3 &
  \ding{55} &
  \ding{55} &
  0.267 &
  0.198 &
  0.581 &
  12.489 &
  0.593 &
  3049.596 \\ \hline
Our FS-Diff &
  \ding{51} &
  \ding{51} &
  \textcolor{top1}{0.737} &
  \textcolor{top1}{0.669} &
  \textcolor{top1}{1.285} &
  \textcolor{top1}{16.170} &
  \textcolor{top1}{0.229} &
  \textcolor{top1}{2496.989} &
  Our FS-Diff &
  \ding{51} &
  \ding{51} &
  \textcolor{top1}{0.533} &
  \textcolor{top1}{0.516} &
  \textcolor{top1}{1.332} &
  \textcolor{top1}{14.716} &
  \textcolor{top1}{0.211} &
  \textcolor{top1}{2370.454} \\ \hline
 &
  \multicolumn{2}{l}{Method} &
  \multicolumn{6}{c|}{MSRS (VI blurred) scale:8} &
   &
  \multicolumn{2}{l}{Method} &
  \multicolumn{6}{c}{AVMS (VI blurred) scale:8} \\ \cline{2-3} \cline{11-12}
 &
  \del{CA-CLIP}\add{CLSE} &
  BFM &
  VIF↑ &
  QAB/F↑ &
  SSIM↑ &
  PSNR↑ &
  LPIPS↓ &
  MSE↓ &
   &
  \del{CA-CLIP}\add{CLSE} &
  BFM &
  VIF↑ &
  QAB/F↑ &
  SSIM↑ &
  PSNR↑ &
  LPIPS↓ &
  MSE↓ \\ \hline
1 &
  \ding{51} &
  \ding{55} &
  0.456 &
  0.567 &
  1.199 &
  15.567 &
  0.339 &
  2949.577 &
  1 &
  \ding{51} &
  \ding{55} &
  0.513 &
  0.445 &
  1.213 &
  9.592 &
  0.322 &
  4365.434 \\
2 &
  \ding{55} &
  \ding{51} &
  0.623 &
  0.534 &
  1.159 &
  14.533 &
  0.317 &
  2694.419 &
  2 &
  \ding{55} &
  \ding{51} &
  0.482 &
  0.442 &
  1.151 &
  11.192 &
  0.426 &
  5423.461 \\
3 &
  \ding{55} &
  \ding{55} &
  0.328 &
  0.598 &
  1.012 &
  12.962 &
  0.511 &
  3359.489 &
  3 &
  \ding{55} &
  \ding{55} &
  0.449 &
  0.297 &
  1.048 &
  10.495 &
  0.311 &
  5119.189 \\ \hline
Our FS-Diff &
  \ding{51} &
  \ding{51} &
  \textcolor{top1}{0.771} &
  \textcolor{top1}{0.683} &
  \textcolor{top1}{1.288} &
  \textcolor{top1}{16.172} &
  \textcolor{top1}{0.229} &
  \textcolor{top1}{2500.538} &
  Our FS-Diff &
  \ding{51} &
  \ding{51} &
  \textcolor{top1}{0.546} &
  \textcolor{top1}{0.523} &
  \textcolor{top1}{1.336} &
  \textcolor{top1}{14.744} &
  \textcolor{top1}{0.211} &
  \textcolor{top1}{2374.510} \\ \hline
 &
  \multicolumn{2}{l}{Method} &
  \multicolumn{6}{c|}{MSRS (IR blurred) scale:8} &
   &
  \multicolumn{2}{l}{Method} &
  \multicolumn{6}{c}{AVMS (IR blurred) scale:8} \\ \cline{2-3} \cline{11-12}
 &
  \del{CA-CLIP}\add{CLSE} &
  BFM &
  VIF↑ &
  QAB/F↑ &
  SSIM↑ &
  PSNR↑ &
  LPIPS↓ &
  MSE↓ &
   &
  \del{CA-CLIP}\add{CLSE} &
  BFM &
  VIF↑ &
  QAB/F↑ &
  SSIM↑ &
  PSNR↑ &
  LPIPS↓ &
  MSE↓ \\ \hline
1 &
  \ding{51} &
  \ding{55} &
  0.412 &
  0.583 &
  1.009 &
  15.277 &
  0.399 &
  2919.504 &
  1 &
  \ding{51} &
  \ding{55} &
  0.466 &
  0.479 &
  1.304 &
  13.489 &
  0.298 &
  2548.192 \\
2 &
  \ding{55} &
  \ding{51} &
  0.647 &
  0.611 &
  1.068 &
  15.192 &
  0.308 &
  2715.189 &
  2 &
  \ding{55} &
  \ding{51} &
  0.492 &
  0.433 &
  1.239 &
  13.994 &
  0.331 &
  2615.919 \\
3 &
  \ding{55} &
  \ding{55} &
  0.348 &
  0.556 &
  0.963 &
  13.489 &
  0.392 &
  3000.489 &
  3 &
  \ding{55} &
  \ding{55} &
  0.348 &
  0.358 &
  1.095 &
  13.496 &
  0.469 &
  2948.953 \\ \hline
Our FS-Diff &
  \ding{51} &
  \ding{51} &
  \textcolor{top1}{0.761} &
  \textcolor{top1}{0.678} &
  \textcolor{top1}{1.287} &
  \textcolor{top1}{16.176} &
  \textcolor{top1}{0.229} &
  \textcolor{top1}{2500.053} &
  Our FS-Diff &
  \ding{51} &
  \ding{51} &
  \textcolor{top1}{0.543} &
  \textcolor{top1}{0.522} &
  \textcolor{top1}{1.335} &
  \textcolor{top1}{14.736} &
  \textcolor{top1}{0.211} &
  \textcolor{top1}{2373.633} \\ \hline
\end{tabular}%
}
\end{table*}

\begin{table}[ht]
\centering
\caption{The impact of content embeddings of different image modalities on fusion performance. (\textcolor{red}{Red}: the best)}
\begin{tabular}{lllllll}
\hline
\label{tab:e}
\multirow{2}{*}{Models} & \multicolumn{6}{c}{Dataset: AVMS fusion dataset (scale:8)} \\

\multicolumn{1}{c}{}              & VIF↑  & QAB/F↑ & SSIM↑ & PSRN↑  & LPIPS↓   & MSE↓  \\\cline{2-7} 
VI clear + IR semantics           & 0.384 & 0.301  & 1.150 & 14.127 & 2377.841 & 0.419 \\
VI clear + VI semantics           &\textcolor{top1}{ 0.543} & \textcolor{top1}{0.522}  & \textcolor{top1}{1.335} & \textcolor{top1}{14.736} & \textcolor{top1}{2373.633} & \textcolor{top1}{0.211} \\\cline{1-7}
IR clear + VI semantics           & 0.385 & 0.303  & 1.153 & 14.145 & 2378.165 & 0.422 \\
IR clear + IR semantics           & \textcolor{top1}{0.546} & \textcolor{top1}{0.523}  & \textcolor{top1}{1.336} & \textcolor{top1}{14.744} & \textcolor{top1}{2374.510} & \textcolor{top1}{0.211} \\\cline{1-7}
Blurred IR \&VI + IR semantics    & 0.423 & 0.365  & 1.217 & 14.331 & 2371.607 & 0.354 \\
Blurred IR \&VI + VI semantics    & 0.414 & 0.350  & 1.204 & 14.270 & 2371.986 & 0.362 \\
Blurred IR \&VI + joint semantics & 0.384 & 0.301  & 1.151 & 14.143 & 2377.909 & 0.421 \\
Blurred IR \&VI + max semantics   & \textcolor{top1}{0.533} & \textcolor{top1}{0.516}  & \textcolor{top1}{1.332} & \textcolor{top1}{14.716} & \textcolor{top1}{2370.454} &\textcolor{top1}{ 0.211}\\ \hline
\end{tabular}
\end{table}

\begin{table}[ht]
\centering
\caption{The impact of CLSE's clarity judgment on the unseen foggy and low resolution AVMS Dataset. (\textcolor{red}{Red}: the best)}
\begin{tabular}{lllllll}
\hline
\label{tab:zero}
\multirow{2}{*}{Structure of CLSE} & \multicolumn{6}{c}{Dataset: AVMS fusion dataset (foggy \& LR (scale:8))} \\
                       & VIF↑      & QAB/F↑     & SSIM↑     & PSRN↑      & LPIPS↓      & MSE↓   \\  \cline{2-7} 
w/o clarity judgment    & 0.379     & 0.293      & 1.105     & 9.920      & 5908.992    & 0.375    \\
w VGG19    &0.348&0.257&1.081&10.684&4902.755&0.434\\
w clarity judgment      & \textcolor{top1}{0.508}     & \textcolor{top1}{0.497}      &\textcolor{top1}{1.319}     & \textcolor{top1}{13.486}     &\textcolor{top1}{2943.330}    & \textcolor{top1}{0.218}   \\ \hline
\end{tabular}
\end{table}

\begin{figure*}[htbp]
\vspace{-0.1cm}
\centering	
\includegraphics[width=1.0\linewidth]{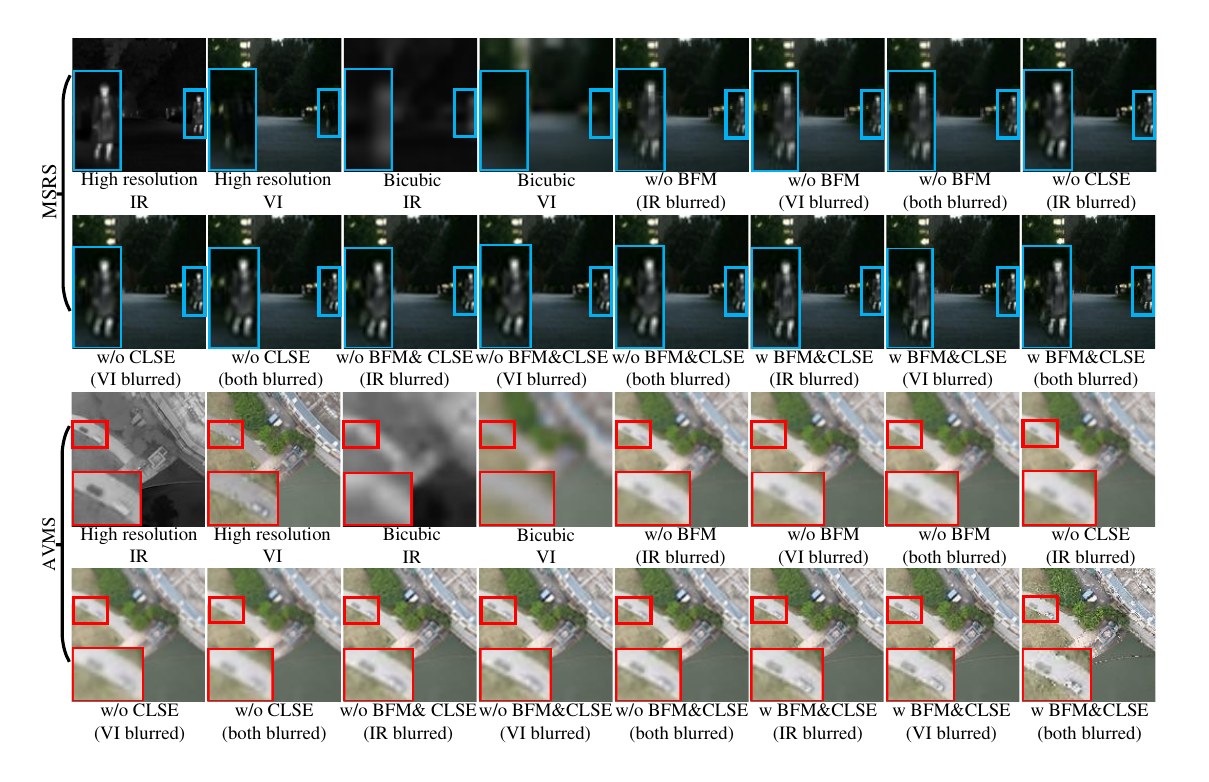}
\caption{Visual comparison of ablation studies on CA-CLIP and CLSE Mechanism.}
\label{Framexr}
\vspace{-0.5cm} % 减少图片下方的垂直间距
\end{figure*}

To investigate the effectiveness of BFM and the CLSE mechanism in our proposed method, we conducted an ablation study on the VIRF task with MSRS and AVMS datasets. The results are presented in Table \ref{tab:3} \aug{and Figure \ref{Framexr}.} BFM plays a crucial role in enabling the joint super-resolution and fusion network to perceive and extract global information from different modal images. The CLSE mechanism, on the other hand, can adaptively sense image clarity and extract semantic information from the input images, \aug{effectively} guiding the fusion process\rem{effectively}. To evaluate their contributions, we removed the BFM and replaced it with a convolutional layer. Additionally, we directly eliminated the CLSE mechanism to assess its impact on semantic guidance during the fusion process.

The ablation experimental results presented in Table \ref{tab:3} \aug{and Figure \ref{Framexr}} demonstrate that the optimal fusion performance is achieved through the synergistic effects of BFM and \del{CA-CLIP}\add{the CLSE mechanism}, thereby validating the rationale behind our FS-Diff framework.

\subsubsection{The discussion of semantic selection in CLSE mechanism}

\begin{figure*}[htbp]
\centering	
\vspace{-0.3cm}
\includegraphics[width=1.0\linewidth]{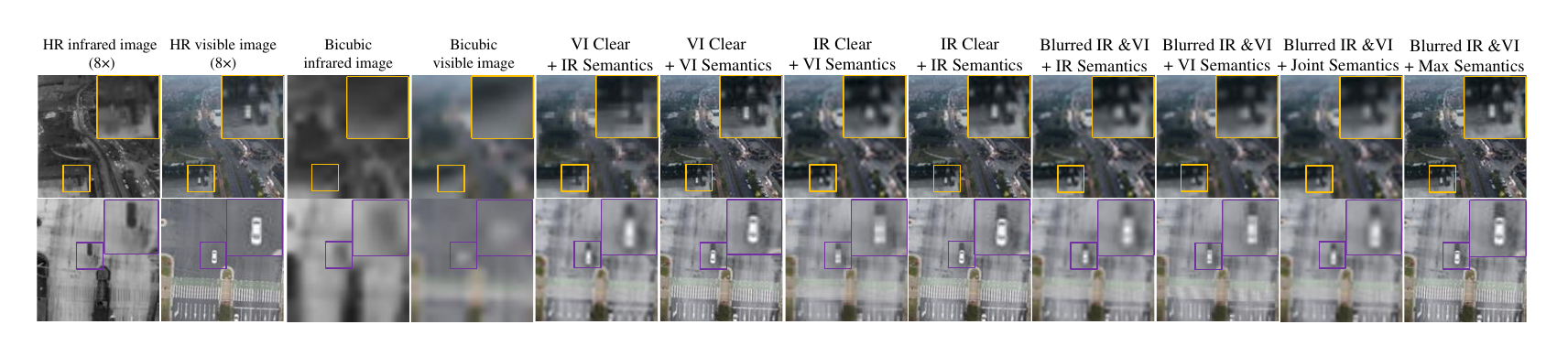}
\caption{The impact of content embeddings from images in different modalities for the fusion results.}
\label{Frameem}
\vspace{-0.3cm} % 减少图片下方的垂直间距
\end{figure*}

\aug{We further corroborate the rationality of the CLSE adaptive feature selection rule design. Specifically, when a single-modal image is blurred, we incorporate the blurred and clear semantics into the fusion process separately. When both modalities are blurred, we embed the blurred single-modal semantics, joint modality semantics, and the maximum of the two-modal semantics into the fusion process, respectively. As shown in Table \ref{tab:e} and Figure \ref{Frameem}, for a single blurred image, choosing the clear semantics and, for two blurred images, selecting the maximum two-modal semantics can effectively enhance fusion performance.}

\subsubsection{The discussion of zero-shot generation of CLSE mechanism}

\begin{figure*}[htbp]
\centering	
\includegraphics[width=1.0\linewidth]{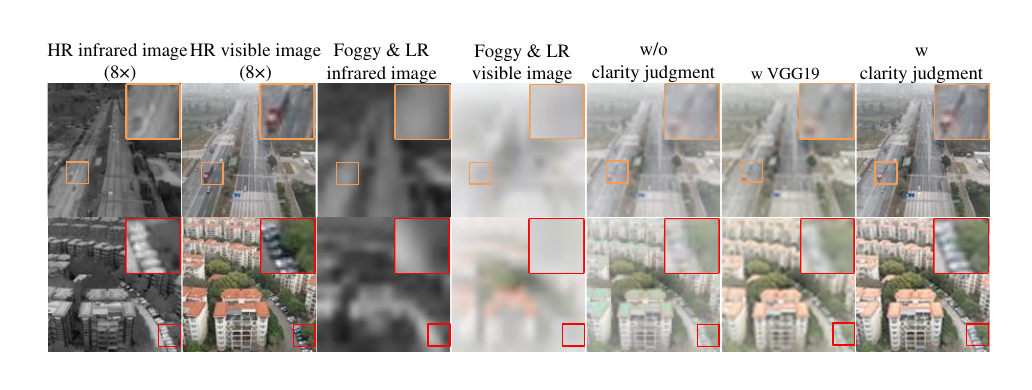}
\caption{Visualization of the effect of CLSE's clarity judgment on the unseen foggy and low resolution (scale:8) AVMS dataset.}
\label{Framezero}
\vspace{-0.3cm}
\end{figure*}

\aug{To evaluate the zero-shot generalization of the adaptive clarity judgment of CLSE, we conducted experiments on the AVMS dataset under compound low-resolution and foggy conditions. In all three experiments, the models did not undergo any prior training and were directly evaluated through inference. As illustrated in Table \ref{tab:zero} and Figure \ref{Framezero}, under the unseen weather conditions, the model integrated with the clarity judgment mechanism exhibits superior performance compared to the model lacking this mechanism and the model substituted with the pre-trained VGG19 network. This cross-condition performance improvement, achieved without specific training, underscores the necessity of the clarity judgment of CLSE.}

\subsection{Performance on advanced visual task}
\label{5.5}
To validate the flexibility of FS-Diff in advanced vision tasks, we conducted experiments on two tasks, namely, image segmentation and detection, using public datasets containing information about larger targets and our AVMS dataset, which was enriched with small targets.

\begin{figure*}[htb]
\centering	
\includegraphics[width=1.0\linewidth]{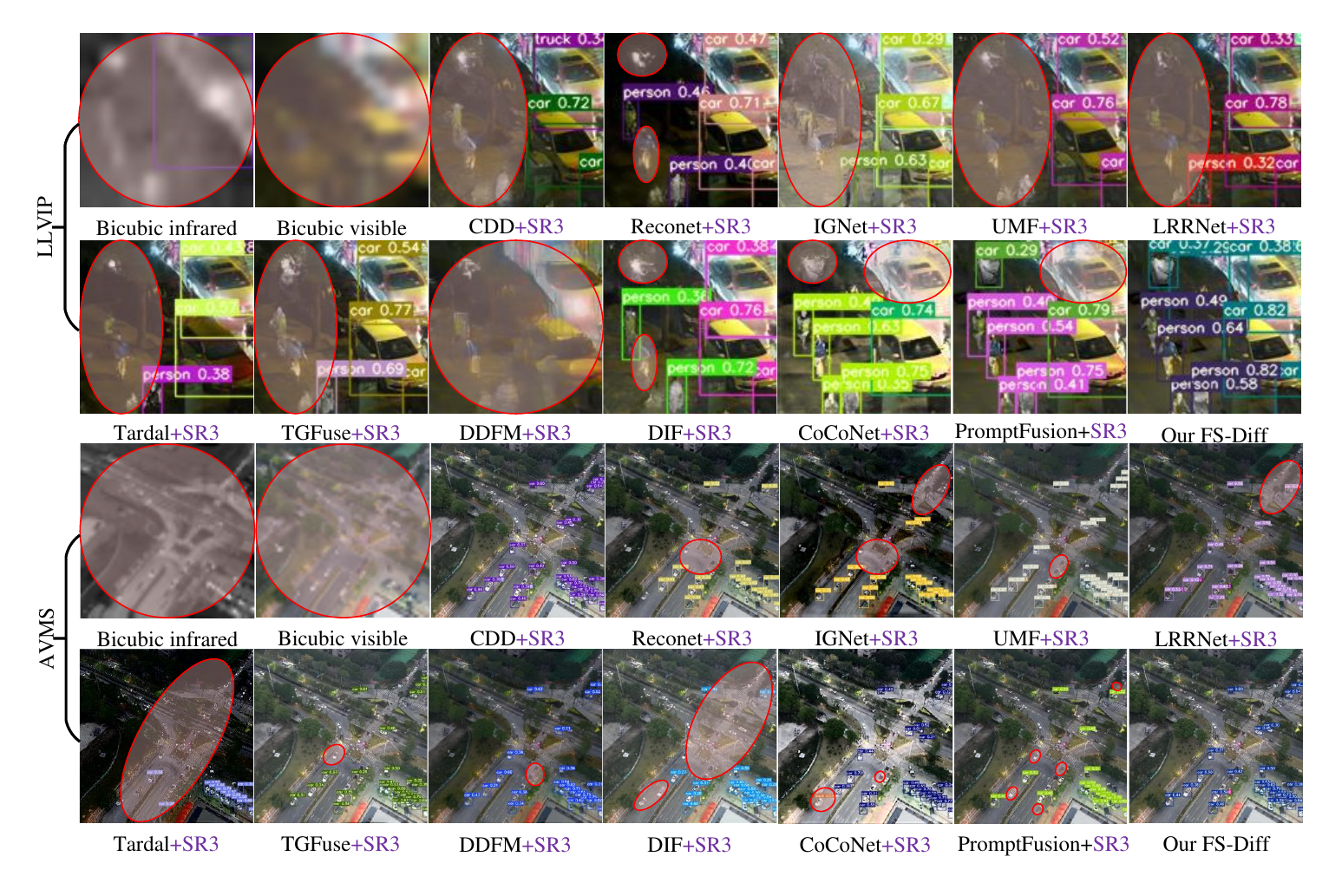}
\caption{Qualitative comparisons of different fusion and super-resolution (scale: 8) detection results on the LLVIP and AVMS datasets. The area circled by the red ellipse indicates the missed detection of the comparison methods compared with our FS-Diff.}
\label{Frame8}
\end{figure*}

\subsubsection{Comparisons on object detection}

% \usepackage{multirow}
% \usepackage{graphicx}
Target detection experiments were conducted on the LLVIP and AVMS datasets, via pretrained YOLOv8 \cite{YOLOv8} as the backbone. Figure \ref{Frame8} shows the results of the qualitative analysis. All the targets in the unimodal images were missed, and mis-detected regions were noted in the comparison methods. By contrast, FS-Diff shows superior detection for the targets of "people" and "cars", obtaining a high confidence score, owing to its rich semantic features. For quantitative analysis, from Table \ref{tab:4}, FS-Diff shows the best detection performance for mAP@0.5, mAP@0.75, and mAP@0.5:0.95, \rem{and it was }followed closely by LRRNet.
\begin{table}[H]
\centering
\caption{Quantitative results of objective detection on LLVIP and AVMS datasets. (\textcolor{red}{Red}: the best, \textcolor{blue}{Blue}: the second best)}
\label{tab:4}
{%
\begin{tabular}{cccc|ccc}
\hline
\multirow{2}{*}{Methods} & \multicolumn{3}{c|}{Dataset:   LLVIP} & \multicolumn{3}{c}{Dataset:   AVMS} \\
                         & mAP@0.50  & mAP@0.75  & mAP@0.50:0.95 & mAP@0.50 & mAP@0.75 & mAP@0.50:0.95 \\ \hline
CDD\textcolor{top4}{+SR3}     & 0.832 & 0.633 & 0.581 & 0.750 & 0.680 & 0.493 \\
Reconet\textcolor{top4}{+SR3} & 0.886 & 0.671 & 0.594 & 0.773 & 0.622 & 0.510 \\
IGNet\textcolor{top4}{+SR3}   & 0.899 & 0.684 & 0.578 & \textcolor{top2}{0.783} & 0.508 & 0.488 \\
UMF\textcolor{top4}{+SR3}     & 0.895 & 0.658 & 0.586 & 0.782 & 0.614 & \textcolor{top2}{0.580} \\
LRRNet\textcolor{top4}{+SR3}  & \textcolor{top2}{0.927} & \textcolor{top2}{0.716} & 0.604 & 0.733 & \textcolor{top2}{0.685} & 0.573 \\
Tardal\textcolor{top4}{+SR3}  & 0.893 & 0.692 & \textcolor{top2}{0.609} & 0.773 & 0.627 & 0.545 \\
TGFuse\textcolor{top4}{+SR3}  & 0.913 & 0.702 & 0.592 & 0.756 & 0.617 & 0.447 \\
DDFM\textcolor{top4}{+SR3}    & 0.851 & 0.623 & 0.515 & 0.740 & 0.522 & 0.462 \\
DIF\textcolor{top4}{+SR3}     & 0.882 & 0.653 & 0.532 & 0.723 & 0.647 & 0.440 \\
CoCoNet\textcolor{top4}{+SR3} & 0.871 & 0.638 & 0.548 & 0.762 & 0.638 & 0.491 \\
PromptFusion\textcolor{top4}{+SR3} & 0.868 & 0.693 & 0.551 & 0.772 & 0.651 & 0.487 \\
Our FS-Diff & \textcolor{top1}{0.933} & \textcolor{top1}{0.720} & \textcolor{top1}{0.624} & \textcolor{top1}{0.795} & \textcolor{top1}{0.728} & \textcolor{top1}{0.598} \\ \hline
\end{tabular}%
}
\end{table}
\subsubsection{Comparisons on segmentation task}
\begin{figure*}[htb]
\centering	
\includegraphics[width=1.0\linewidth]{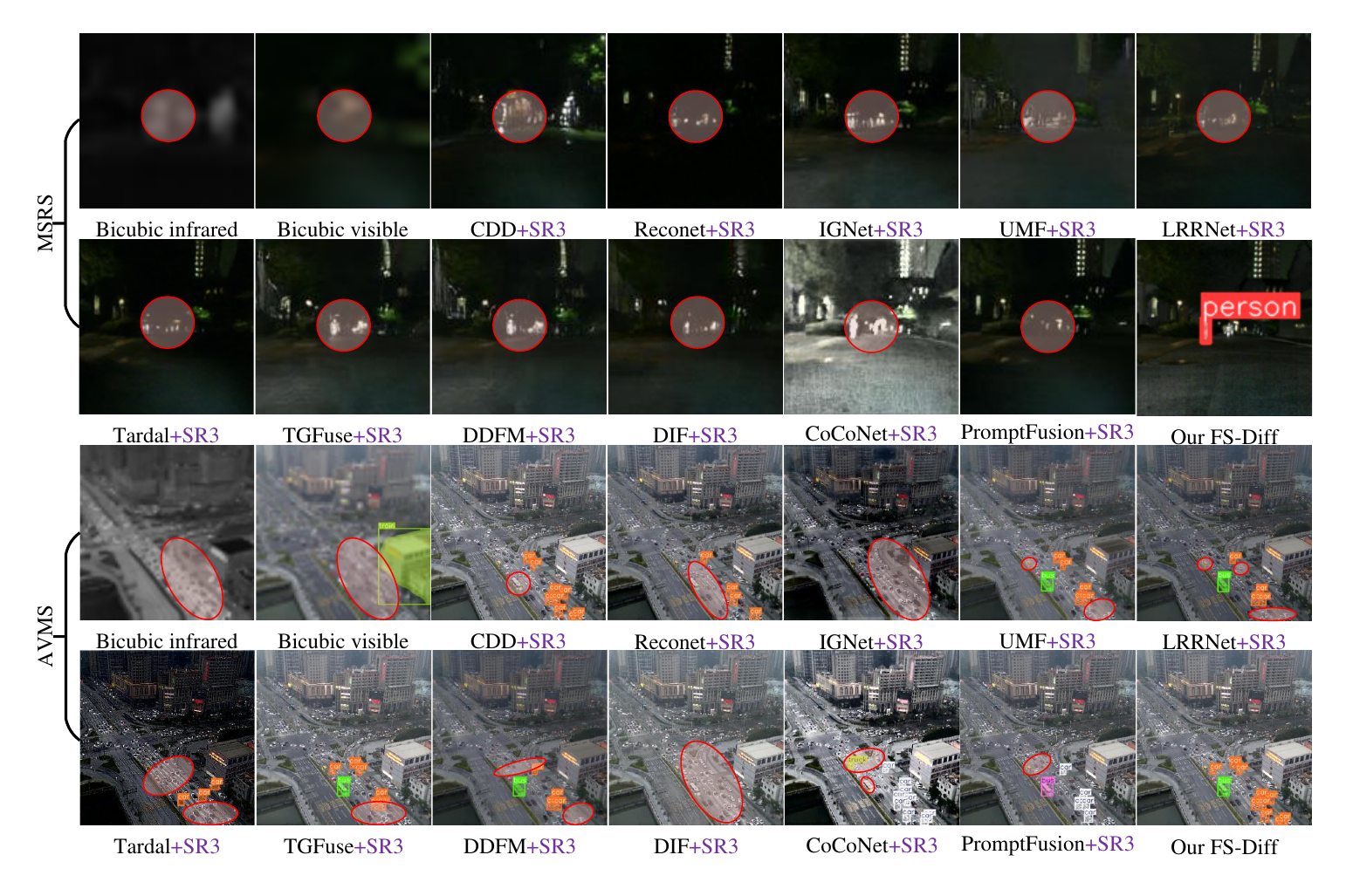}
\caption{Qualitative comparisons of different fusion and super-resolution (scale: 8) segmentation results on the MSRS and AVMS datasets. The red regions represent the missing segmentations of the comparison methods compared with our FS-Diff.}
\label{Frame9}
\end{figure*}
Image segmentation experiments were conducted on the MSRS and our AVMS datasets, using BiSeNet \cite{94} retrained on the MSRS as the backbone of the segmentation task. The qualitative analysis shown in Figure \ref{Frame9} shows that the segmentation results of unimodal LR images are generally worse than the fusion results. \rem{In addition, due to the loss of the semantic information of the images after fusion and super-resolution as well as the accumulation of error information, the comparison methods appear to lose segmentation regions. By contrast to our FS-Diff segmentation results, other methods have difficulty in accurately segmenting “people” in MSRS, which suggests that FS-Diff can highlight hot target information.} \aug{In addition, the separate processes of fusion and super-resolution lead to the loss of semantic information and the accumulation of errors, causing the comparison methods to fail in preserving segmentation regions. In contrast, FS-Diff produces more accurate segmentation results. For example, in the MSRS dataset, other methods struggle to \rem{accurately} segment "people" \aug{accurately}, whereas FS-Diff effectively highlights the hot target information.} Moreover, CDD, Reconet, IGNet, Tardal, DIF \aug{and CoCoNet} have difficulty in accurately segmenting “buses” in AVMS. Although FS-Diff fails to completely segment all vehicles on the road, it segments the most semantic information, compared with other methods.

The quantitative analyses are shown in Table \ref{tab:5}. The mIoU values of FS-Diff on the MSRS and AVMS datasets were 3.596\% and 2.151\% higher than those of the second-ranked DDFM and IGNet, respectively. These results show that FS-Diff exhibits superior performance in image segmentation tasks.
\begin{table}[H]
\centering
\caption{Quantitative results of segmentation on MSRS and AVMS datasets. (\textcolor{red}{Red}: the best, \textcolor{blue}{Blue}: the second best)}
\label{tab:5}
\resizebox{\columnwidth}{!}{%
\begin{tabular}{lllllllllll}
\hline
 &
  \multicolumn{10}{c}{Dataset: MSRS} \\
\multirow{-2}{*}{Methods} &
  Background &
  Car &
  Person &
  Bike &
  Curve &
  Car stop &
  Guardrail &
  Color tone &
  Bump &
  mIoU \\ \hline
CDD\textcolor{top4}{+SR3} &
  93.057 &
  79.195 &
  69.879 &
  62.939 &
  50.165 &
  64.259 &
  79.475 &
  67.937 &
  71.707 &
  0.710 \\
Reconet\textcolor{top4}{+SR3} &
  92.856 &
  67.922 &
  70.125 &
  69.354 &
  57.234 &
  54.216 &
  78.244 &
  \textcolor{top2}{68.129} &
  57.242 &
  0.684 \\
IGNet\textcolor{top4}{+SR3} &
  92.852 &
  78.684 &
  77.337 &
  68.555 &
  \textcolor{top2}{63.552} &
  62.729 &
  \textcolor{top1}{83.242} &
  66.843 &
  43.686 &
  0.708 \\
UMF\textcolor{top4}{+SR3} &
  92.73 &
  74.861 &
  71.165 &
  56.371 &
  51.462 &
  51.451 &
  60.465 &
  62.771 &
  63.729 &
  0.650 \\
LRRNet\textcolor{top4}{+SR3} &
  92.953 &
  79.078 &
  71.545 &
  \textcolor{top2}{71.291} &
  63.462 &
  \textcolor{top2}{64.574} &
  77.251 &
  67.375 &
  53.353 &
  0.712 \\
Tardal\textcolor{top4}{+SR3} &
  92.589 &
  62.474 &
  70.087 &
  65.239 &
  61.368 &
  61.577 &
  78.465 &
  66.033 &
  72.467 &
  0.700 \\
TGFuse\textcolor{top4}{+SR3} &
  93.093 &
  \textcolor{top2}{80.447} &
  69.735 &
  62.331 &
  54.327 &
  55.088 &
  72.461 &
  66.305 &
  \textcolor{top2}{72.614} &
  0.696 \\
DDFM\textcolor{top4}{+SR3} &
  \textcolor{top1}{93.551} &
  78.251 &
  75.259 &
  69.267 &
  58.247 &
  62.457 &
  78.366 &
  64.251 &
  71.250 &
  \textcolor{top2}{0.723} \\
DIF\textcolor{top4}{+SR3} &
  92.493 &
  72.117 &
  60.644 &
  64.827 &
  48.428 &
  61.537 &
  73.496 &
  59.242 &
  62.591 &
  0.662 \\
CoCoNet\textcolor{top4}{+SR3} &
  92.489 &
  79.415 &
 \textcolor{top1} {79.546} &
  68.367 &
  62.189 &
  51.159 &
  71.156 &
  61.524 &
  67.128 &
  0.699 \\
PromptFusion\textcolor{top4}{+SR3} &
  93.195 &
  72.156 &
  72.235 &
  \textcolor{top1}{76.951} &
  57.462 &
  57.416 &
  78.954 &
  66.554 &
  71.089 &
  0.717\\  
Our   FS-Diff &
  \textcolor{top2}{93.217} &
  \textcolor{top1}{81.561} &
  \textcolor{top2}{77.341} &
  69.846 &
  \textcolor{top1}{63.627} &
  \textcolor{top1}{64.968} &
  \textcolor{top2}{82.523} &
  \textcolor{top1}{68.951} &
  \textcolor{top1}{72.961} &
  \textcolor{top1}{0.749} \\ \hline
 &
  \multicolumn{10}{c}{Dataset: AVMS} \\
\multirow{-2}{*}{Methods} &
  Background &
  Car &
  Person &
  Bike &
  Truck &
  %Electric bicycle&
  \begin{tabular}[c]{@{}l@{}}Electric \\ bicycle\end{tabular} &
  Tent &
  Grab &
  Bus &
  mloU \\ \hline
CDD\textcolor{top4}{+SR3} &
  \textcolor{top1}{0.980} &
  0.734 &
  0.803 &
  0.769 &
  0.585 &
  0.445 &
  0.609 &
  0.156 &
  \textcolor{top2}{0.718} &
  0.644 \\
Reconet\textcolor{top4}{+SR3} &
  0.946 &
  0.719 &
  0.773 &
  0.711 &
  0.577 &
  \textcolor{top2}{0.654} &
  0.557 &
  0.167 &
  0.682 &
  0.643 \\
IGNet\textcolor{top4}{+SR3} &
  \textcolor{top2}{0.979} &
  0.689 &
  \textcolor{top1}{0.889} &
  0.768 &
  0.489 &
  0.585 &
  \textcolor{top2}{0.660} &
  0.106 &
  0.693 &
  \textcolor{top2}{0.651} \\
UMF\textcolor{top4}{+SR3} &
  0.907 &
  \textcolor{top2}{0.865} &
  0.667 &
  0.714 &
  \textcolor{top1}{0.596} &
  0.522 &
  0.519 &
  0.111 &
  0.698 &
  0.622 \\
LRRNet\textcolor{top4}{+SR3} &
  0.972 &
  0.844 &
  0.675 &
  0.742 &
  0.571 &
  0.437 &
  \textcolor{top1}{0.693} &
  0.138 &
  0.627 &
  0.633 \\
Tardal\textcolor{top4}{+SR3} &
  0.947 &
  0.864 &
  0.778 &
  0.530 &
  0.572 &
  0.642 &
  0.404 &
  0.104 &
  0.506 &
  0.594 \\
TGFuse\textcolor{top4}{+SR3} &
  0.905 &
  0.774 &
  0.772 &
  \textcolor{top2}{0.776} &
  \textcolor{top1}{0.596} &
  0.571 &
  0.508 &
  0.178 &
  0.590 &
  0.630 \\
DDFM\textcolor{top4}{+SR3} &
  0.967 &
  0.854 &
  0.681 &
  0.610 &
  0.529 &
  \textcolor{top1}{0.660} &
  0.536 &
  0.116 &
  0.650 &
  0.623 \\
DIF\textcolor{top4}{+SR3} &
  0.933 &
  0.810 &
  0.821 &
  0.631 &
  0.441 &
  0.449 &
  0.617 &
  0.123 &
  \textcolor{top1}{0.787} &
  0.623 \\
CoCoNet\textcolor{top4}{+SR3} &
  0.956 &
  0.824 &
  0.862 &
  0.751 &
  0.492 &
  0.427 &
  0.546 &
  \textcolor{top2}{0.246} &
  0.613 &
  0.635 \\
PromptFusion\textcolor{top4}{+SR3} &
  0.948 &
  0.831 &
  0.857 &
  0.734 &
  0.490 &
  0.413 &
  0.551 &
  \textcolor{top1}{0.257} &
  0.603 &
  0.632 \\
Our   FS-Diff &
  0.976 &
  \textcolor{top1}{0.869} &
  \textcolor{top2}{0.875} &
  \textcolor{top1}{0.824} &
  0.511 &
  0.438 &
  0.657 &
  0.223 &
  0.613 &
  \textcolor{top1}{0.665} \\ \hline
\end{tabular}%
}
\end{table}

\subsection{Extended experiments}
\label{5.6}
\begin{table}[H]
\centering
\caption{The objective results for multi-focus image fusion and super resolution (\textcolor{red}{Red}: the best, \textcolor{top2}{Blue}: the second best)}
\label{tab:6}
{%
\begin{tabular}{lllllll}
\hline
\multirow{2}{*}{Methods} & \multicolumn{6}{c}{Dataset: Lytro   (scale:8)}     \\
                         & VIF↑  & Q\textsuperscript{AB/F}↑ & SSIM↑ & PSNR↑  & LPIPS↓ & MSE↓    \\ \hline
FusionDiff\textcolor{top4}{+SR3}           & 0.974 & 0.648  & 1.642 & 22.387 & 0.114  & 742.285 \\
RRSTD\textcolor{top4}{+SR3}                & \textcolor{top2}{1.136} & \textcolor{top2}{0.744}  & \textcolor{top2}{1.787} & \textcolor{top2}{27.449} & \textcolor{top2}{0.076}  & 205.217 \\
MUFusion\textcolor{top4}{+SR3}             & 1.112 & 0.735  & 1.784 & 25.360 & 0.077  & 240.048 \\
U2Fusion\textcolor{top4}{+SR3}             & 1.030 & 0.723  & 1.755 & 23.683 & 0.080  & 338.688 \\
SR3-only(far)             & 1.091 & 0.723 &  1.783 &
\textcolor{top1}{27.635} & 0.078  &\textcolor{top1}{192.624}\\
SR3-only(near)          & 1.112 &0.734 &\textcolor{top2}{1.787} &\textcolor{top1}{27.635} &\textcolor{top2}{0.076}  &\textcolor{top2}{198.685}\\
Our   FS-Diff            & \textcolor{top1}{1.269} & \textcolor{top1}{0.779}  & \textcolor{top1}{1.797} & 27.384 & \textcolor{top1}{0.074}  & 221.278 \\ \hline
\multirow{2}{*}{Methods} & \multicolumn{6}{c}{Dataset: MFI   (scale:8)}       \\
                         & VIF↑  & Q\textsuperscript{AB/F}↑ & SSIM↑ & PSNR↑  & LPIPS↓ & MSE↓    \\ \hline
FusionDiff\textcolor{top4}{+SR3}           & 0.965 & 0.635  & 1.681 & 21.979 & 0.102  & 713.968 \\
RRSTD\textcolor{top4}{+SR3}                & \textcolor{top2}{1.365} & \textcolor{top2}{0.832}  & \textcolor{top2}{1.937} & 33.401 & \textcolor{top2}{0.029}  & 30.820  \\
MUFusion\textcolor{top4}{+SR3}             & 1.344 & 0.827  & 1.928 & 27.444 & 0.031  & 122.392 \\
U2Fusion\textcolor{top4}{+SR3}             & 1.267 & 0.819  & 1.913 & 26.634 & 0.033  & 172.695 \\
SR3-only(far)             & 1.362 &0.831  &\textcolor{top2}{1.937} & 
\textcolor{top1}{33.436} & \textcolor{top2}{0.029} & \textcolor{top1}{30.578} \\
SR3-only(near)        &1.361 &0.831 &\textcolor{top2}{1.937} &\textcolor{top2}{33.420} &\textcolor{top2}{0.029} &\textcolor{top2}{30.612}  \\
Our   FS-Diff            & \textcolor{top1}{1.442} & \textcolor{top1}{0.855}  & \textcolor{top1}{1.941} & {32.266} & \textcolor{top1}{0.028}  & {60.797}  \\ \hline
\end{tabular}%
}
\end{table}

\begin{figure*}[htb]
\centering	
\includegraphics[width=1.0\linewidth]{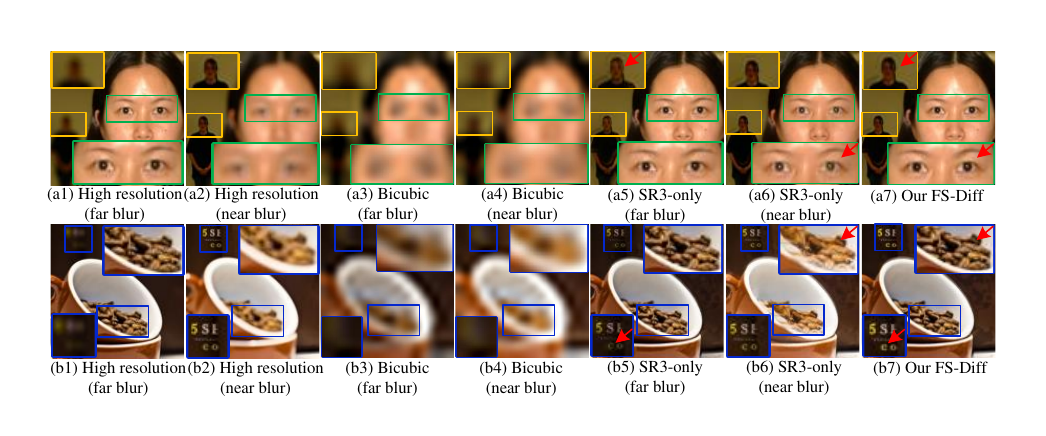}
\caption{
Visual Comparison of SR3-only super-resolution and the proposed method for simultaneous multi-source image fusion and super-resolution (scale:8).
}
\label{Framefocus}
\end{figure*}

\begin{figure*}[htb]
\centering	
\includegraphics[width=1.0\linewidth]{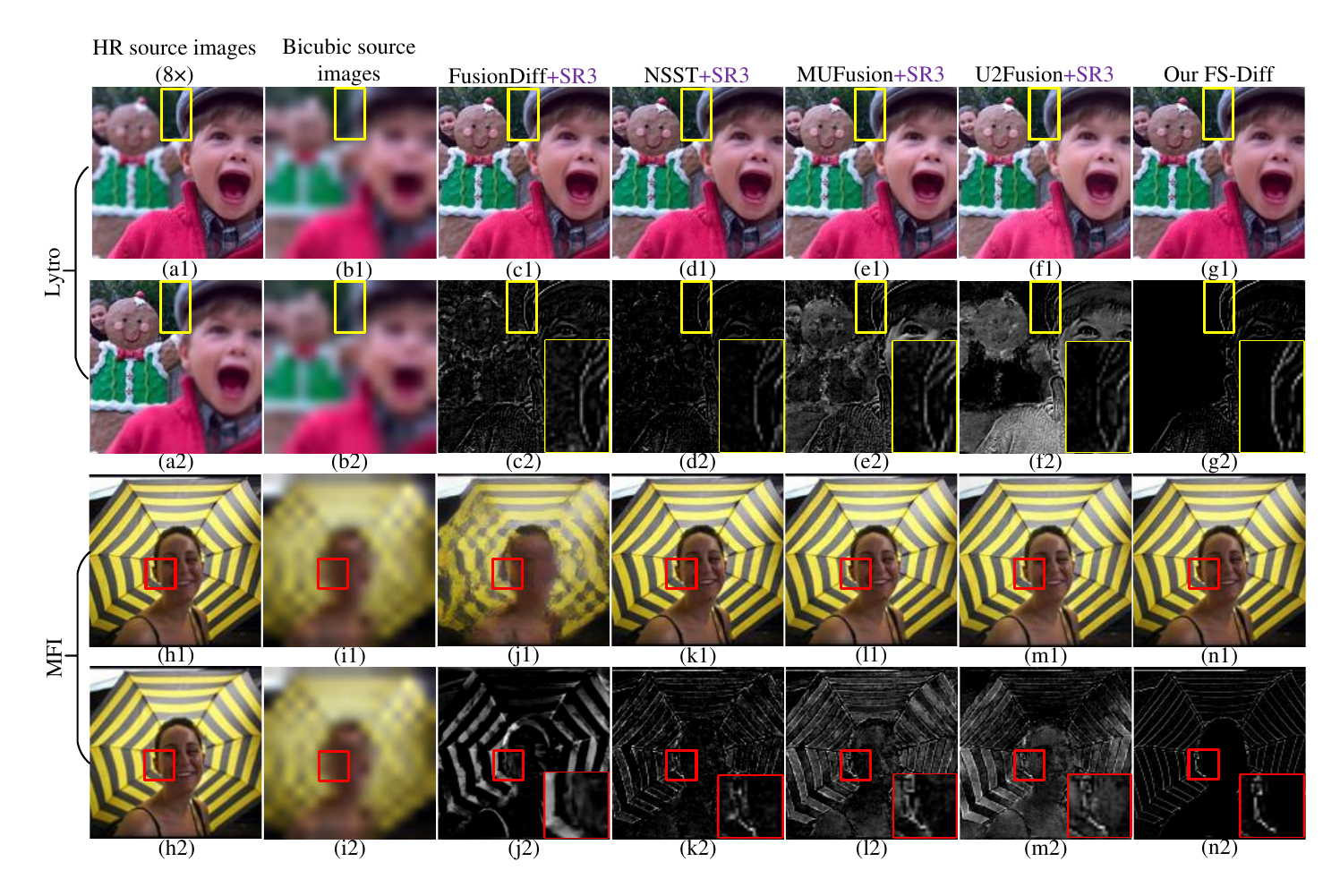}
\caption{Fusion and super-resolution (scale: 8) results for different methods on the Lytro and MFI multi-focus datasets. (a1)–(a2) and (h1)–(h2) are partially focused source images; (b1)–(b2) and (i1)–(i2) are input images obtained via bicubic downsampling; (c1)–(g1) and (j1)–(n1) are fused results; (c2)–(g2) are the differences produced by (c1)–(g1) with (a2); and (j2)–(n2) are the differences produced by (j1)–(n1) with (h2).}
\label{Frame10}
\end{figure*}

\aug{Existing multi-focus image fusion methods \cite{97,29,98,19,95,96} are limited to fixed resolutions. Although strategies such as performing super-resolution prior to fusion can be employed, the former may introduce artifacts that, when amplified in subsequent steps, significantly degrade the final result. Simultaneous super-resolution and fusion of multifocus images enhance the information content in the fused result, preserving more focused features from the source images.} To further validate the generalizability of FS-Diff, we compared it with FusionDiff \cite{97}, RRSTD \cite{29}, MUFusion \cite{98}, and U2Fusion \cite{19} on the public datasets Lytro \cite{95} and MFI \cite{96}. The stepwise fusion and super-resolution methods were used for comparison. For training, HR fusion images were obtained via U2Fusion \cite{19} and training image pairs were obtained by bicubically downsampling the source images. To be consistent with the above experiment, we centered and cropped the two datasets, degraded them to \(16 \times 16\), and termed them Lytro-val and MFI-val, respectively.

The quantitative evaluation under the condition where both image
modalities are blurred in Table \ref{tab:6} shows that FS-Diff achieved the best values for the VIF, Q\textsuperscript{AB/F}, SSIM and LPIPS metrics, with RRSTD closely following. Compared \rem{to}\aug{with} the SR3-only (far/near) method, which applies super-resolution to the two source images, FS-Diff has significant advantages in preserving visual information fidelity, structural similarity, fusion quality, and \rem{perceptual}\aug{visual} consistency. \aug{As shown in Figure \ref{Framefocus}, despite the super-resolution processing of the low-resolution image (Figures \ref{Framefocus} (a3)-(a4), 
(b3)-(b4)), the out-of-focus information remains arduous to recover (Figures \ref{Framefocus} (a5)-(a6), (b5)-(b6)).} For a more intuitive visual evaluation, Figures \ref{Frame10} (c2)-(g2) and (j8)-(n2) show the differences obtained using the different methods. Less information in the difference map is obtained by subtracting the focus area in one source image from the fusion image, implying that more focus information is extracted by the fusion method. \rem{As can be seen from the}\aug{The} enlarged area in the lower right corner of Figure \ref{Frame10}, FusionDiff, MUFusion, NSST, and U2Fusion have different levels of residuals around the edge of the hat of the “child” and around the ears of the “lady with the umbrella.” In addition, FusionDiff’s stepwise fusion and super-resolution for the “lady with the umbrella” is unsatisfactory, with a \rem{serious}\aug{severe} loss of semantic information. By contrast, Figures \ref{Frame10} (g2) and (n2) show the most continuous and cleanest hat and ear edges, respectively, and preserve the continuous boundary information. 

\rem{The above quantitative and qualitative evaluations show that FS-Diff has excellent performance and good model generalization for multi-focus image fusion.}

%Although SR3-only achieves higher PSNR and lower MSE, it falls short in retaining visual and structural fidelity.

% \usepackage{multirow}
% \usepackage{graphicx}
%\add{
\section{Computational complexity analysis}

%\begin{table}[htbp]  % 尝试将表格放置在合适的位置
%  \centering  % 使表格居中
%  \caption{Computational efficiency and model complexity analysis}
%  \label{tab:7}
%  \resizebox{\textwidth}{!}{%
%   \begin{tabular}{llllllllllllll}  % 定义一个有三列的表格
%    \toprule
%    Methods       & CDD   & Reconet & IGNet & UMF   & LRRNet                                                      & Tardal & TGFuse & DDFM    & DIF   & MetaFusion & FS-Diff & Diff-IF & CoCoNet  \\  % 表头
%    \midrule
%    Parameters(M) & 1.19  & $7.50 \times 10^{-3}$  & 7.87 & 0.63  & 755.57                                                      & 0.30   & 137.34 & 552.81  & 32.20 & 0.75       & 58.47 & 23.71 &9.13    \\  % 第一行数据
%    FLOPs(G) & 29.21 & 1.52    & 13.49 & 10.30 & $4.87 \times 10^{-5}$ &38.89  & 3.99   & 1114.71 & 14.97 & 97.18      & 64,391.60 &43.53 &10.39 \\  % 第二行数据
%    Time(s) &0.0167 &  0.0021   & 0.005 & 0.0019 & 0.0088 & 0.087  &  0.013 &34.50  & 0.072 &  0.077     &  74&0.29&0.015  \\ 

%    \bottomrule
%  \end{tabular}
%  }
%\end{table}

\begin{table}[htbp]
  \centering  % 使表格居中
  \caption{Comparison of computational complexity and inference time}
  \label{tab:7}
  \resizebox{\textwidth}{!}{%
\begin{tabular}{@{}llllllll@{}}
\toprule
Methods       & CDD\textcolor{top4}{+SR3}        & Reconet\textcolor{top4}{+SR3}   & IGNet\textcolor{top4}{+SR3}     & UMF\textcolor{top4}{+SR3}      & LRRNet\textcolor{top4}{+SR3}   & Tardal\textcolor{top4}{+SR3} & TGFuse\textcolor{top4}{+SR3} \\ \midrule
Parameters(M) & 1.19\textcolor{top4}{+97.76}      & 0.01\textcolor{top4}{+97.76}  & 7.87\textcolor{top4}{+97.76}    & 0.63\textcolor{top4}{+97.76}    & 755.57\textcolor{top4}{+97.76}     & 0.30\textcolor{top4}{+97.76}    & 137.34\textcolor{top4}{+97.76} \\
FLOPs(G)      & 29.21\textcolor{top4}{+91681.6}   & 1.52\textcolor{top4}{+91681.6}  & 13.49\textcolor{top4}{+91681.6} & 10.30\textcolor{top4}{+91681.6} & $4.90\times10^{-5}$
\textcolor{top4}{+91681.6} & 38.89\textcolor{top4}{+91681.6} & 3.99\textcolor{top4}{+91681.6} \\
Time(s)       & 0.017\textcolor{top4}{+$n$×35}     & 0.002\textcolor{top4}{+$n$×35}   & 0.005\textcolor{top4}{+$n$×35}    & 0.002\textcolor{top4}{+$n$×35}   & 0.009\textcolor{top4}{+$n$×35}      & 0.087\textcolor{top4}{+$n$×35}    & 0.013\textcolor{top4}{+$n$×35}   \\ \bottomrule
Methods       & DDFM\textcolor{top4}{+SR3}       & DIF\textcolor{top4}{+SR3}       & Diff-IF\textcolor{top4}{+SR3}   & CoCoNet\textcolor{top4}{+SR3}  & PromptFusion\textcolor{top4}{+SR3} & MetaFusion & FS-Diff    \\ \midrule
Parameters(M) & 552.81\textcolor{top4}{+97.76} & 32.20\textcolor{top4}{+97.76} & 23.71\textcolor{top4}{+97.76} & 9.13\textcolor{top4}{+97.76} &    7.44\textcolor{top4}{+97.76}        & 0.75       & 58.47      \\
FLOPs(G)      & 1114.71\textcolor{top4}{+91681.6} & 14.97\textcolor{top4}{+91681.6} & 43.53\textcolor{top4}{+91681.6} & 10.39\textcolor{top4}{+91681.6} & 44.56\textcolor{top4}{+91681.6}                 & 97.18           & 64,391.60      \\
Time(s)       & 34.50\textcolor{top4}{+$n$×35}   & 0.072\textcolor{top4}{+$n$×35}  & 0.290\textcolor{top4}{+$n$×35}   & 0.015\textcolor{top4}{+$n$×35} &   0.189\textcolor{top4}{+$n$×35}           & 0.077      & 74.0         \\ \bottomrule
\end{tabular}
}
\end{table}

\aug{
This section compares the computational efficiency and memory consumption between our proposed method and the compared methods. The FLOPs and parameters are quantified at the input size of 128×128, \rem{while}\aug{whereas} the running time is averaged over ten 128×128 images from the AVMS dataset.}
From Table \ref{tab:7}, the proposed FS-Diff method has \rem{the parameters}\aug{parameter} of 58.47M, which is slightly \rem{higher}\aug{greater} than methods like Reconet, UMF and Tardal, but still moderate compared to LRRNet and DDFM. In terms of computational complexity, FS-Diff \rem{achieves 64,391.6G FLOPs, }significantly higher than CDD and Reconet, primarily \rem{due to}\aug{because of} the simultaneous execution of image super-resolution and fusion tasks. Despite the increased computational cost and inference time, FS-Diff delivers finer image details and superior fusion quality. Furthermore, when adding the SR3 parameters\rem{ (97.7597M)}, the FLOPs \rem{(91,681.6G)} \aug{and inference time (where $n=1$ when a single image is blurred and $n=2$ when both images are blurred)} are used for separately performing super-resolution and fusion tasks for the compared methods, the overall computational burden far exceeds that of FS-Diff.

\section{Conclusion}
To address the resolution damage to image targets and backgrounds caused by long-distance shooting and the limitations of single imaging devices in fully capturing scene information \rem{due to}\aug{because of} sensor constraints, we proposed FS-Diff, a unified image super-resolution and fusion framework. By presenting the \add{CLSE mechanism with} CA-CLIP, FS-Diff achieves adaptive \del{resolution}\add{clarity} awareness for both single- and dual-source images with reduced resolution\aug{.} \rem{ and}\aug{It also} provides semantic guidance during the fusion process. Additionally, we introduced the BFM to \del{construct a joint representation of inputs, enhancing} \add{enhance} the \add{global} feature extraction of \del{cross-modal information for multi-source}\add{multimodal} images. Taking LR multimodal images and semantics guidance as conditions, we generated HR-fused results through an iterative refinement denoising process in an end-to-end manner, effectively preserving the high-fidelity detail and cross-modal information required for high-magnification fused images. We conducted extensive experiments on three image fusion tasks: VIRF, MIF, and multifocus image fusion. The results of the quantitative and qualitative evaluations fully demonstrated the strong robustness of FS-Diff in joint fusion and super-resolution tasks with multiple high-magnification ratios as well as its application value in advanced visual tasks. In addition, to broaden the application boundaries of image fusion methods from an aerial photography perspective, we constructed an AVMS dataset\aug{.} \rem{ that included}\aug{It includes} a variety of lighting intensities\aug{,} \rem{and}scenes, \rem{ that provided}\aug{and} 3821 annotations for objects. This dataset can be used for \del{MSIF}\add{MMIF}, image super-resolution, object detection, and semantic segmentation tasks, providing a new benchmark for research in computer vision.

Although the FS-Diff framework demonstrated significant effects on joint image super-resolution and fusion, it has limitations such as high model complexity and long running times. In the future, we plan to \add{improve the U-Net structure of our model to reduce model complexity and}\del{combine an external memory module to capture sample-specific attributes of the training dataset and guide image generation during the testing phase, which will reduce the number of computational steps and accelerate the inference process to} further enhance the ability of FS-Diff to \rem{quickly} adapt \aug{quickly} to complex and variable real-world scenarios. Simultaneously, we are integrating further research on FS-Diff into autonomous driving systems, UAV precision agricultural irrigation technology, and UAV military combat strategies\aug{.} \aug{This will}\rem{to} expand its applications and practices in various fields, including intelligent transportation, agricultural modernization, and defense science and technology.

\section{Acknowledgment}
This research was supported by the National Natural Science Foundation of China (No. 62201149), the Natural Science Foundation of Guangdong Province (No. 2024A1515011880), the Basic and Applied Basic Research of Guangdong Province (Nos. 2023A1515140077, 2024A1515012287), the Science and Technology Key Program of Guangzhou (No. 2023B03J1388), and the National Key R\&D Pogram of China (No. 2023YFA1011601).

 \bibliographystyle{elsarticle-num} 
 \bibliography{ref}




% %% For numbered reference style

%   \textit{\LaTeX: a document preparation system},
%   Addison Wesley, Massachusetts,

% \end{thebibliography}
\end{document}